\documentclass[pdflatex,sn-mathphys-num]{sn-jnl}

\usepackage{lmodern} 
\usepackage{graphicx}
\usepackage{booktabs}
\usepackage{multirow}
\usepackage{algorithm}
\usepackage[noend]{algpseudocode}
\usepackage{xspace}
\usepackage{amsmath,amssymb,amsthm,amsthm,thmtools,thm-restate,bm,stmaryrd}
\usepackage{adjustbox}
\usepackage{enumitem}
\usepackage{chngcntr}

\usepackage{titlesec}
\titleformat{\paragraph}[runin]{\normalfont\normalsize\bfseries}{}{1em}{}
\titlespacing*{\paragraph}{0pt}{1.25ex plus 1ex minus .2ex}{1em}

\def\bbleft{\llbracket}
\def\bbright{\rrbracket}

\usepackage{tikz}
\usetikzlibrary{calc}

\usepackage{caption}

\usepackage{cleveref}
\usepackage{graphicx}
\usepackage{booktabs}
\usepackage{multirow}

\theoremstyle{theorem}

\newtheorem{lemma}{Lemma}
\newtheorem{proposition}{Proposition}
\theoremstyle{remark}

\theoremstyle{definition}
\newtheorem{example}{Example}
\crefname{assumption}{Assumption}{Assumptions}
\crefname{figure}{Figure}{Figures}
\crefname{table}{Table}{Tables}
\crefname{section}{Section}{Sections}
\crefname{appendix}{Appendix}{Appendices}
\crefname{theorem}{Theorem}{Theorems}
\crefname{lemma}{Lemma}{Lemmas}
\crefname{example}{Example}{Examples}
\crefname{proposition}{Proposition}{Propositions}
\crefformat{footnote}{#2\footnotemark[#1]#3}
\newcommand{\red}{\color[rgb]{1,0,0}}

\def\threshold{\tau}
\def\I{\mathcal{I}}
\def\M{\mathcal{M}}
\def\N{\mathcal{N}}
\def\T{^{\mathsf T}}
\def\Real{\mathbb{R}}

\DeclareMathOperator*{\argmax}{arg\,max}
\DeclareMathOperator*{\argmin}{arg\,min}
\DeclareMathOperator*{\sigm}{sigm}

\def\l1{\ensuremath{\ell_1}\xspace}
\def\l2{\ensuremath{\ell_2}\xspace}

\newcommand{\diag}{\operatorname{diag}}

\newcommand{\smax}{\operatorname{smax}}

\newcommand{\vk}{\mathbf{k}}

\newcommand{\vw}{\mathbf{w}}
\newcommand{\vx}{\mathbf{x}}

\makeatletter
\DeclareRobustCommand\onedot{\futurelet\@let@token\@onedot}
\def\@onedot{\ifx\@let@token.\else.\null\fi\xspace}
\def\eg{\emph{e.g}\onedot} 
\def\ie{\emph{i.e}\onedot} 
\def\cf{\emph{cf}\onedot} 
\def\etc{\emph{etc}\onedot} 
\def\wrt{w.r.t\onedot}

\makeatother

\def\mid{\,|\,}

\usepackage[absolute,overlay]{textpos}  

\begin{document}

\begin{textblock*}{\paperwidth}(0cm,1cm) 
  \centering
  \large\textbf{Preprint version, submitted to IJCV, December 2025}
\end{textblock*}


\def\mytitle{RANSAC Scoring Functions: \\ Analysis and Reality Check}
\title{\mytitle}


\author*{\fnm{Alexander} \sur{Shekhovtsov}}\email{shekhole@fel.cvut.cz}



\affil{\orgdiv{Visual Recognition Group, Department of Cybernetics}, \orgname{Czech Technical University in Prague}, \orgaddress{\street{Karlovo náměstí 13}, \city{Prague}, \postcode{121 35}, \country{Czech Republic}}}


\abstract{
We revisit the problem of assigning a score (a quality of fit) to candidate geometric models --- one of the key components of RANSAC for robust geometric fitting.
In a non-robust setting, the ``gold standard'' scoring function, known as the geometric error, follows from a probabilistic model with Gaussian noises. We extend it to spherical noises. In a robust setting, we consider a mixture with uniformly distributed outliers and show that a threshold-based parameterization leads to a unified view of likelihood-based and robust M-estimators and associated local optimization schemes.

Next we analyze MAGSAC++ which stands out for two reasons. First, it achieves the best results according to existing benchmarks. Second, it makes quite different modeling assumptions and derivation steps. We discovered, however that the derivation does not correspond to sound principles and the resulting score function is in fact numerically equivalent to a simple Gaussian-uniform likelihood, a basic model within the proposed framework.

Finally, we propose an experimental methodology for evaluating scoring functions: assuming either a large validation set, or a small random validation set in expectation. We find that all scoring functions, including using a learned inlier distribution, perform identically. In particular, MAGSAC++ score is found to be neither better performing than simple contenders nor less sensitive to the choice of the threshold hyperparameter.

Our theoretical and experimental analysis thus comprehensively revisit the state-of-the-art, which is critical for any future research seeking to improve the methods or apply them to other robust fitting problems.
}




\maketitle

\section{Introduction}
\label{sec:intro}
RANdom SAmple Consensus (RANSAC)~\cite{RANSAC} is a well-known method for robust estimation of geometric models. Example applications include finding geometric primitives (lines, circles, planes, \etc), relative pose estimation from correspondences in two or more views, camera localization from 3D-to-2D correspondences, panoramic stitching, rigid motion, and other. If a minimal number of noise-free correspondences were known, the model could be found by solving respective polynomial equations. \citet{Bhayani-24} give an overview of such minimal solvers. When the correspondences have noisy coordinates but there are sufficiently many of them one can refine the model by continuous optimization~\cite{Chojnacki-00}.
When the correspondences are both noisy and contaminated by false matches, RANSAC and robust optimization are applied. RANSAC randomly generates subsets of correspondences and applies minimal solvers.
The problem is thus reduced to selecting the best model from a finite set of generated candidates and refining the best model found. For both the best model selection within RANSAC and model refinement via continuous robust optimization, a quality criterion is necessary, which we will call a {\em score}. A robust score, which accounts for unknown outliers and utilizes the inliers statistically optimally, is a key component of robust and accurate estimation. 

The score considered in the basic version of RANSAC is the number of ``inliers'' --- correspondences with a geometric error within a specified tolerance threshold. It is clearly not suitable for continuous refinement, because it discards fine-grained information. During a long history of RANSAC, many improved scores have been proposed, inspired by robust estimators (MSAC~\cite{Torr-98}), probabilistic models (MLSAC~\cite{torr2002bayesian}), or a combination of both (MAGSAC\cite{barath2019magsac}, MAGSAC++\cite{barath2021marginalizing}).
The focus of this work is to theoretically understand and compare engineered score functions, as well as to verify the prior experimental evidence regarding their performance and to study the limitations.

First, we revisit a commonly used probabilistic model for non-robust and robust estimation, explicitly discussing the assumptions and showing how they lead to a simplified 1D mixture model of residuals measured with Sampson error.
We then establish a correspondence between the maximum likelihood (ML) and several robust estimators. Toward this end, we show that the simple mixture model with uniform outlier residuals gives rise to robust scoring functions through profile or marginal likelihood formulations.
Parameterizing the model by the decision threshold parameter allows us to simplify scoring functions as well as associated fitting algorithms and compare them directly. In particular, we show how Expectation Maximization (EM), popular in machine learning, corresponds to Iterative Reweighted Least Squares (IRLS), popular in robust statistics. This is presented in \cref{sec:model,sec:m-estimators}.

Second, we analyze MAGSAC/MAGSAC++, which is widely accepted as a state-of-the-art (SOTA) method for relative pose and homography estimation problems: it is included as part of OpenCV library, according to~\citet{ivashechkin2021vsac} it is the most accurate RANSAC variant, it is believed to be more robust and less sensitive to the choice of the threshold thanks to marginalization over scales~\cite{barath2021marginalizing}. Its main innovation is a robust M-estimator and the associated score, based on modeling inlier residuals with chi-distribution, scale-marginalization and IRLS refinement.
If there are serious issues with the correctness and validity of the results (and there are), it is of utmost importance to analyze them --- to identify their causes and understand consequences. 
This is crucial for researchers to advance the methods and for practitioners to make informed decisions about extending the methods to other domains.
This detailed analysis, based on the background developed in \cref{sec:model,sec:m-estimators}, is the subject of \cref{sec:M++}.

Finally, in the experimental evaluation in \cref{sec:experiments} we compare different scoring functions head-to-head, \ie within exactly the same basic pipeline. To achieve an objective comparison with respect to the choice of hyperparameters, we propose two approaches: either using a sufficiently large validation set so that hyperparameters can be determined reliably or considering the results in expectation over a small random validation set. The verification, conducted for homographies, essential and fundamental matrices, disproves many previously assumed indirect experimental evidence. First, there is no significant improvement of any scoring method tested over the classical MSAC scoring, neither for minimal model selection nor for local optimization. In particular, the improvements in~\cite{barath2021marginalizing, barath2022learning} must be due to factors other than the scoring function. Second, the sensitivities of all the methods to the choice of the threshold hyperparameter are far more similar than previously measured. Additionally, we disprove the experimental evidence~\cite{barath2022learning} that more accurate models may have, on average, fewer inliers than the less accurate ones, which was the main motivation for using a general neural network for model scoring~\cite{barath2022learning}.

We conclude with a discussion of the limitations of residual-based scoring and possible ways of overcoming them.

\section{Related Work}\label{sec:related}

\subsection{Prior Theoretical Analysis}
While probabilistic models have been considered for RANSAC for a long time (\eg~\cite{torr2000mlesac}), the underlying assumptions and the steps necessary to reduce the likelihood to a function of residuals are often only briefly discussed. Similarly, while robust estimators are motivated by a particular distribution of inliers, \eg Gaussian for MSAC~\cite{torr2000mlesac} or chi for MAGSAC++, these motivations are not formalized, which leads, as we will show, to critical conceptual errors. 

\subsection{Prior Evaluation}
Many works have evaluated different RANSAC variants~\cite{Jin-21,ivashechkin2021vsac,Barath_2023_CVPR,barath2021marginalizing}. However, they evaluate {\em complete pipelines}. The experiments suggest that some scoring functions, in particular MAGSAC++, being the core of the MAGSAC++ pipeline, perform substantially better than others. However, the pipelines differ in more aspects than just the score, therefore no clear conclusions regarding scoring functions can be drawn. We build an independent testing framework targeting specifically evaluating scoring functions and show, using a refined evaluation methodology, that there is no objective improvement.

\citet[Fig.2]{Barroso-23} conducted experimental analysis of the reasons for RANSAC failure. They highlight that suboptimal candidate model selection due to scoring is responsible for a significant percentage of failure cases, especially for more complex datasets. They address this problem with an image-based score, while we stick with the geometry-based score. For the future work it is promissing to combine both.

\subsection{Alternative Scoring Methods}
This study is limited to the most widely used scoring functions based on residuals. However, there are many alternative approaches. Instead of threshold selection, RANSAAC~\cite{rais2017accurate} takes a weighted average of the estimated models, each represented as a set of points.
MINPRAN~\cite{stewart1995minpran} assumes uniform noise and adaptively sets the threshold, minimizing the randomness of the points that fall close to the model. A contrario RANSAC~\cite{moulon2012adaptive} minimizes the expected number of false alarm points, \ie, the number of inconsistent points within a specific threshold. 
\citet{torr2000mlesac,Heinrich-13} and more recently \citet{Edstedt_2025_CVPR} estimate the inlier scale per image pair from the data, with the latter methods shown to be bennefficial in practice.

\citet{Brachmann_2017_CVPR}, \citet{barath2022learning} and \citet{Barroso-23} have considered learned quality functions using deep neural networks, claiming substantial improvements in SOTA and attracting attention to this aspect of RANSAC. However, neural network solutions require substantially more computations, which limits their applicability for minimal model scoring, and even more so for robust refitting through continuous optimization.
The quality function of \citet{Barroso-23} works on the image level. It is more robust but also more expensive to compute and less precise. 
The improvements demonstrated by \citet{barath2022learning} are not necessarily due to the score, as they combine the learned score with the inlier count and add a heavy post-processing. Furthermore, the results are not reproducible (neither code nor the network is public). The merit of learning-based scores has not in our opinion been clearly demonstrated.

\subsection{Other Robust Fitting Problems}\label{sec:other-robust}
While we focus on homography / relative pose estimation, the proposed theoretical framework and experimental methodology are in principle applicable to other robust fitting problems, such as absolute pose estimation~\cite{sattler2018benchmarking,sarlin2022lamar}, point cloud registration~\cite{pomerleau2015review} and 3D geometric model fitting.
In particular, in cases where classical RANSAC or MAGSAC++ has been previously applied (\eg plane and cylinder fitting to 3D point clouds~\cite{Ling_2024_ISPRS,Nurunnabi_2017_ISPRS}, PnP~\cite{wang2024spacecraftPoseMAGSAC}, \etc) our analysis and methods would be applicable directly. 
Other domains might present additional challenges, such as different sensors (LiDAR) and noise characteristics.

\section{Scoring by Probabilistic Modeling}\label{sec:model}
We start by discussing a probabilistic model combining inliers and outliers. The probabilistic model has a non-trivial interaction with the geometry, which we will explore in detail.

\subsection{Geometry}
Consider the problem of robust estimation of a geometric model from observations (detections or tentative correspondences) $\{x_i \in \mathbb{R}^d\}_{i=1}^n$. Let us give several common examples.


\paragraph{Relative Pose} The geometric model is the essential matrix $E \in \Real^{3\times 3}$ describing the relative pose of two cameras, an observation $x_i = (u_i, v_i) \in \Real^4$ is a pair of corresponding detections $u,v \in \Real^2$ in the two views. The true correspondence $\bar x = (\bar u, \bar v) \in \Real^4$ must be a projection of a 3D point and therefore must satisfy the scalar equation $(\bar v, 1)^{\sf T} E (\bar u, 1) = 0$,~\cite{hartley2003multiple}.

\paragraph{Homography} The geometric model is a homography matrix $H \in \Real^{3\times 3}$ and the correspondences are as above. True correspondences must satisfy $\bar v_i = \lambda_i H \bar u_i$ for some free $\lambda_i>0$, totaling 2 constraints per correspondence.

\paragraph{Absolute Pose}The geometric model is the pose of the camera in the world coordinates represented by rotation $R$ and translation $t$. Let $X_i$ be a known 3D point, $x_i$ a detection of this point in the camera and $\bar x_i \in \Real^2$ the true projection of $X_i$ to the camera. The true projection $\bar x$ must satisfy 2 equations, \ie it is uniquely defined by $X_i$ and the camera pose.

\begin{figure}[t]
\centering
\setlength{\tabcolsep}{10pt}
\resizebox{0.9\linewidth}{!}{
\begin{tabular}{ccc}
\includegraphics[width=0.29\linewidth]{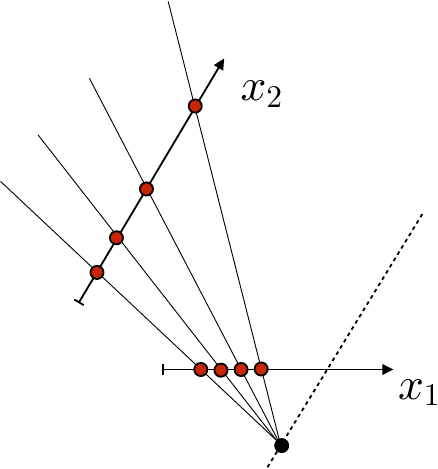} \ \ \ \ \ &
\includegraphics[width=0.28\linewidth]{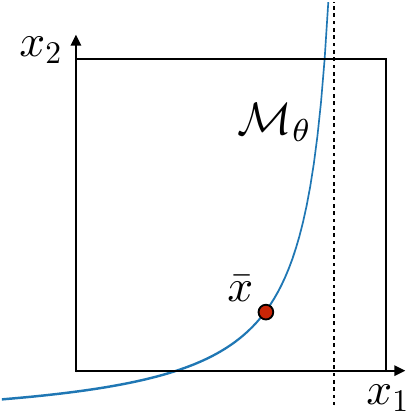}& \ \ \ \ \ \ \ \ \ \  
\includegraphics[width=0.24\linewidth]{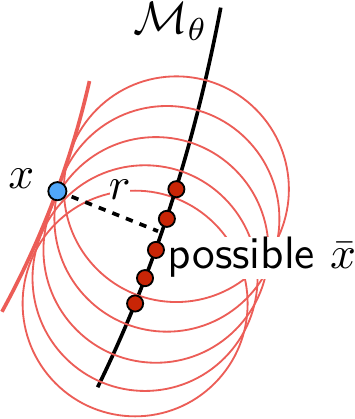}\\
a & b & c
\end{tabular}
}
\caption{(a) Correspondences under homography in a plane. (b) Manifold $\M_{\theta}$ is a hyperbola. 
(c) Knowing the observed correspondence $x$ and marginalizing over true unknown correspondence on the manifold $\bar x$, the resulting probability $p(x; \theta)$ is a function of only the distance from $x$ to the manifold $\M_{\theta}$.
\label{fig:model}}
\end{figure}

In all of the above examples, the true correspondence (or projection) $\bar x \in \Real^d$ must satisfy algebraic constraints $g(\bar x, \theta) = 0$, where $\theta$ denotes the model hypothesis and $g$ is vector-valued in general. Let $\mathcal{M}_\theta = \{x \in \mathbb{R}^d \mid  g(x, \theta) = 0\}$ be the manifold of points that satisfy the constraint. \cref{fig:model} (b) illustrates it for homography. If the function $g$ is $d_g$ dimensional, the manifold dimension is $d_\M  = d - d_g$ ($d_g$ is co-dimension of $\M$).
In a non-robust setting, a common approach is to score $\theta$ by the {\em geometric error}~\cite{Chojnacki-00,hartley2003multiple,Terekhov_2023_ICCV}:
\begin{align}
\mathcal{E}(\theta) = \min_{\stackrel{\bar x}{g(\bar x, \theta) = 0}} (x - \bar x)^2,
\end{align}
which follows from the maximum likelihood for a statistical model with Gaussian noise, treating $\bar x$ as nuisance parameters estimated together with $\theta$. The best model $\theta$, found by minimizing $\mathcal{E}$, can therefore be referred to as the {\em gold standard} maximum likelihood estimate~\cite{hartley2003multiple,Terekhov_2023_ICCV}. A nonspherical Gaussian noise model is considered by \cite{Chojnacki-00}.

\subsection{The Basic Probabilistic Model}
In a robust setting as well, an established statistical approach is to introduce a probabilistic model and choose $\theta$ that maximizes the likelihood of observations in this model. 
Many works quickly jump over to considering a 1D model of residuals, typically taken to be the Sampson errors. We believe it is important to understand which underlying assumptions allow for this transition and which 1D distribution occurs. Detailing the complete model will also help us understand its limitations (\ie to see which assumptions are not realistic) and explain the issues with MAGSAC++.

The probabilistic model we consider is based on the following synthetic generative process. 
Given $\theta$, each correspondence is generated independently.
We first make a random draw to determine whether each correspondence is an {\em inlier} or {\em outlier}. Let $k \in\{0,1\}$ be the respective Bernoulli variable with $p(k{=}1) = \gamma$.
If $k=1$, a true correspondence $\bar x \in \M_\theta$ is generated 
from a uniform or slowly varying distribution on $\M_{\theta}$. Then, $x \in \Real^d$ is generated as a noisy observation of $\bar x$.

\subsubsection{Inliers}
Commonly, the inlier density $p_{\rm in}(x \mid \bar x)$, or the {\em observation noise}, is considered to be Gaussian $\N(\bar x, \sigma^2 I)$ and the true unknown $\bar x$ is considered a nuisance variable to be maximized over jointly with $\theta$~\cite{Chojnacki-00}. 
We will consider a somewhat more general case when the observation noise is {\em spherical}: $p_{\rm in}(x \mid \bar x) = \frac{1}{Z} h(\|x - \bar x \|)$, where $h(\rho)$ is the {\em radial profile function} and $Z$ is the normalization constant.
We assume that on the scale of the observation noise the manifold is approximately linear and the distribution of $\bar x$ is approximately uniform. Furthermore, instead of maximizing, we marginalize out the unknown $\bar x$. The noise density $p_{\rm in}(x \mid \bar x)$ gets sum-projected over $d_\M$ dimensions and the resulting density $p_{\rm in}(x; \theta)$ becomes a function only of the distance $r(x, \theta)$ from $x$ to $\M_\theta$, as illustrated in~\cref{fig:model}(c) for homography. Because $p_{\rm in}(x \mid \bar x)$ is spherical about $\bar x$, the marginal distribution $p_{\rm in}(x; \theta)$ is ``cylindrical'' about the manifold $\M_\theta$: it is spherical over $d_g$ dimensions orthogonal to the manifold and constant along $d_\M$ dimensions spanning the manifold. Therefore we can write the proportionality
\begin{align}\label{ray-density}
p_{\rm in}(x; \theta) \propto p_{\rm in}(r{=}r(x; \theta)),
\end{align}
where $p_{\rm in}(r)$ is the profile function of the marginal density orthogonally to the manifold. It can be interpreted as a density along any ray orthogonal to the manifold, so we refer to it as the {\em ray density}.

\begin{figure}[t]
    \centering
\includegraphics[width=\linewidth]{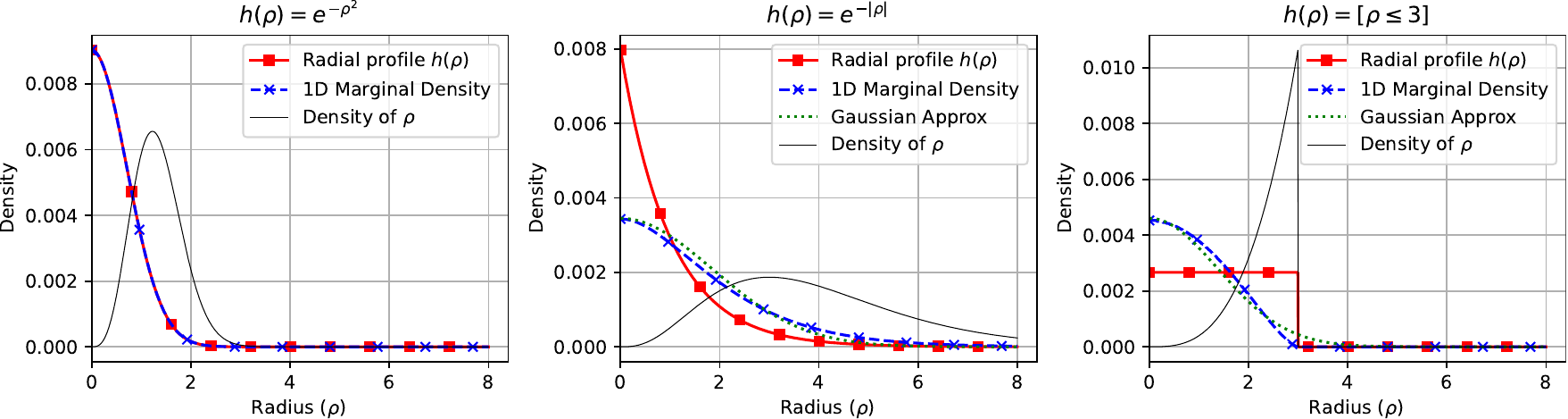}
\caption{Examples of different profile functions for a spherical density in 4D: Gaussian, Laplacian and uniform. Their corresponding 1D marginal distribution (blue), integrating 3 dimensions out of 4 is close to Gaussian in all cases. The distribution of distances $\rho = \|x - \bar x \|$, considered in MAGSAC++ is cordinally different.
    \label{fig:spherical}}
    \centering
\setlength{\tabcolsep}{0pt}
\begin{tabular}{cc}%
\includegraphics[width=0.5\linewidth]{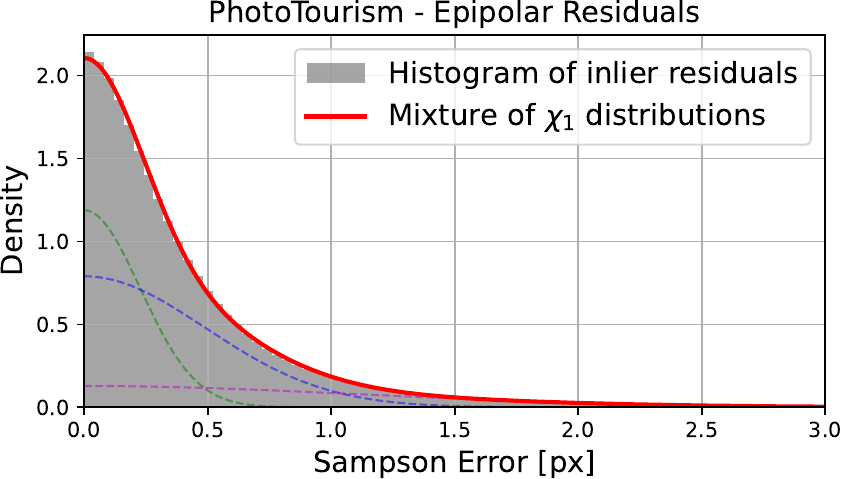}&%
\includegraphics[width=0.5\linewidth]{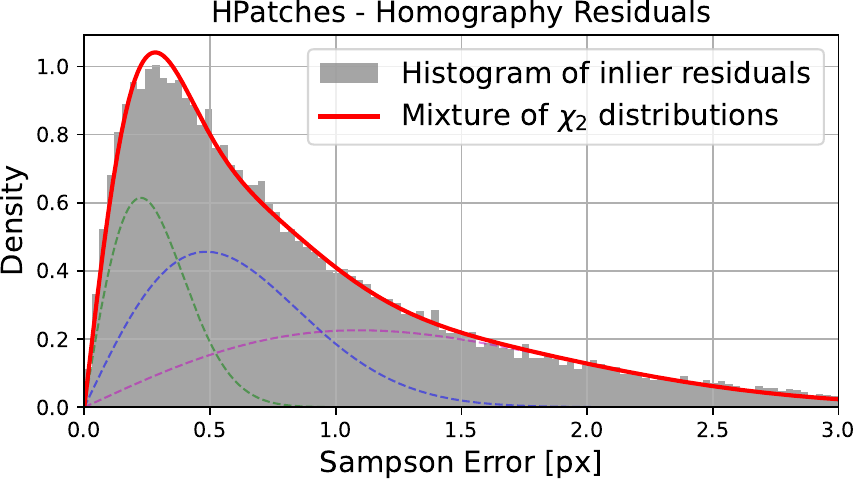}
\end{tabular}
\caption{Statistics of Sampson errors of inliers in real data: {\em left}: essential/fundamental matrix on PhotoTourism ($d_g$ = 1, $\chi_1$ distributed according to the model); {\em right}: homography on HPatches ($d_g = 2$, $\chi_2$ distributed according to the model). Data histograms are showin in gray and a fitted mixture of chi distributions with different scales is shown in red, dashed lines show mixture components.
\label{fig:inliers}}
\end{figure}

\paragraph{Relative Pose}
Assume the observation noise distribution is 4D spherical with a radial profile $h(\rho)$. The manifold $\M_\theta$ has dimension $d_\M = 3$ and co-dimension $d_g=1$. In this case the ray density $p_{\rm in}(r)$  coincides with the marginal distribution of residuals $p(r)$. It is illustrated in \cref{fig:spherical} for different profiles $h(\rho)$ and it is seen that because of marginalization over $d_\M$ dimensions, it tends to normality regardless of $h(\rho)$. We also contrast the ray density $p_{\rm in}(r)$ with the density of the distances $\rho = \| x - \bar x\|$ (\cref{fig:spherical}, black), considered in MAGSAC++. 
If the observation noise is Gaussian, then $\rho$ is chi-distributed with $4$ degrees of freedom, $\chi_4$, while $p_{\rm in}(r)$ is exactly 1D Gaussian\footnote{Here and in similar context, Gaussian is meant to be restricted to the support $[0, \infty]$ and renormalized, which can be seen the same as $\chi_1$ distribution.}. We will see that MAGSAC++ mistakenly confuses these two very different distributions.


\paragraph{Homography}
Assume for simplicity that the observation noise is Gaussian. The marginal distribution $p_{\rm in}(x; \theta)$ for Homography is spherical in 2 dimensions because $d_g=2$. Its ray density $p_{\rm in}(r; \theta)$ is still a 1D Gaussian while the marginal density of residuals $r$ is $\chi_2$. 

In \cref{fig:inliers} we show inlier distributions of Sampson errors for epipolar geometry and homography on real data. The inliers are estimated using GT models from PhotoTourism~\cite{Jin-21} and HPatches~\cite{HPatches} datasets (see~\cref{A:exp-inliers} for more details). Since SIFT keypoints are used for matching, the inlier noise is expected to be close to Gaussian~\cite{Lowe-04}, albeit possibly at different scales (according to image resolution and scale-space processing). This agrees with the fitted mixture distributions.
\citet{Edstedt_2025_CVPR} proposes more experimental evidence of Gaussian-like residual distributions for relative pose in real data with different matchers.




The distance $r(x; \theta)$ from $x$ to $\M_\theta$ can be computed in practice by the Sampson error. It is the length of one Newton step for solving the non-linear equation $g(x,\theta) = 0$ starting from $x$,\footnote{In 2 or higher dimensions, a Gauss-Newton step for minimizing $\|g(x,\theta)\|^2$.}. It is easy to compute and was experimentally verified to be rather accurate, in particular for homography~\cite{Chum-97} and relative pose~\cite{rydell2024revisiting}.

\subsubsection{Outliers and The Model Summary}
It remains to discuss the model for outliers ($k=0$). In this case we assume that the correspondence $x \in \Real^d$ is a random mismatch so that its components are completely unrelated. Hence it is suitable to model $p_{\rm out}(x)$ as uniform over the observation domain. Up to the domain constraints, the uniform density $p_{\rm out}(x)$ is also spherical with a constant ray density $p_{\rm out}(r)$.

We can now summarize the {\em basic probabilistic model} as follows.
Let $\vx = (x_i)_{i=1}^n$ be the vector of correspondences, $\vk = (k_i)_{i=1}^n$ the vector of inlier states, and $r_i(\theta) = r(x_i, \theta) \geq 0 $ the {\em residual} of the correspondence $x_i$ with respect to the model $\theta$. Under the above assumptions, the likelihood of observations simplifies as:
\begin{align}\label{basic-model}
\textstyle p(\vx, \vk; \theta) \propto \prod_{i}\big( \gamma p_{\rm in}(r_i)\bbleft k_i=1 \bbright + (1-\gamma) p_{\rm out}(r_i)\bbleft k_i=0 \bbright \big),
\end{align}
where $p_{\rm in}(r)$ is the 1D ray density of inlier residuals and $p_{\rm out}(r)$ is the 1D ray density of outlier residuals. 
In this simplified model, we may assume that $p_{\rm in}(r)$ has most of its mass in some range $[0, T]$ for a large enough $T$ and that $p_{\rm out}(r)$ is constant, \ie equals some $\alpha < \frac{1}{T}$ for $r \in [0,T]$. Thus we see that the likelihood can indeed be reduced to product of 1D densities. 
It is important not to confuse the 1D {\em ray density} $p_{\rm in}(r)$ with the distribution of residuals to GT positions $\rho = \| x - \bar x\|$, or with the marginal distribution of distances $r$ between $x$ and the manifold, which may take different shapes, as illustrated in~\cref{fig:spherical,fig:inliers}. We have seen that the ray density $p_{\rm in}(r)$ is 1D Gaussian when the observation noise is Gaussian, for any $d_g$, and tends to Gaussian even if the observation noise is not Gaussian. 
The marginal density of residuals $r$, which tends to $\chi_{d_g}$ may be used for a valid test of inlier vs outlier (as \eg in~\cite{hartley2003multiple}), but it is not suitable for defining the likelihood and scoring the candidate models, as will be discussed in~\cref{sec:MAGSAC-likelihood}.


\section{From Likelihood to M-estimators}\label{sec:m-estimators}
\subsection{Marginal and Profile Likelihoods}
For estimating $\theta$ in the basic probabilistic model, we can consider the {\em marginal likelihood} maximization, $\max_\theta \sum_{\vk} p(\vx,\vk; \theta)$, where the inlier state variables $\vk$ are marginalized out, or the {\em profile likelihood} maximization, $\max_\theta \max_{\vk} p(\vx,\vk; \theta)$, where we are profiling over $\vk$ (finding the most likely model $\theta$ and the inlier states $\vk$ jointly). If we are primarily interested in estimating $\theta$ and not $\vk$, the marginal likelihood is more sound, but the profile likelihood is also applicable and may be simpler.

The logarithms of the marginal and profile likelihoods in the basic model define two possible {\em model quality functions (scores)}:  
\begin{subequations}
\begin{align}\label{marg-quality}
    \textstyle Q^{\text{\sc marg}}(\theta) &= \textstyle \log \sum_{\vk} p(\vx,\vk; \theta) = \sum_i \log \Big( \gamma p_{\rm in}(r_i) + (1-\gamma) p_{\rm out}(r_i) \Big),\\
\label{profile-quality}
\textstyle Q^{\text{\sc prof}}(\theta) &= \textstyle \max_{\vk} \log p(\vx,\vk; \theta) = \sum_i \log \max \big( \gamma p_{\rm in}(r_i), (1-\gamma) p_{\rm out}(r_i) \Big).
\end{align}
\end{subequations}
Both expressions take the form $\sum_i \rho(r_i)$ with a corresponding {\em residual scoring function} $\rho(r)$.

\subsection{Likelihood-based M-estimators}
Our next step is to parameterize both types of scoring functions in terms of a decision threshold $\tau$. For profile likelihood, this threshold corresponds to the decision about inliers made in the inner maximization problem. For marginal likelihood, no hard decision will be made; however, adopting the same parameter $\tau$ will allow for a uniform treatment and transparent comparison.

The {\em maximum a posteriori} decision strategy is to classify a point with residual $r$ as an inlier if
$\gamma p_{\rm in}(r) > (1-\gamma) p_{\rm out}(r)$.
Using the uniformity assumption of outliers in $[0,T]$, we can define the set of inliers using a threshold rule on the inlier density:
 \begin{align}
\textstyle \I = \{ i \mid p_{\rm in}(r_i) > \mu \text{\ and \ }  r_i<T \},
\end{align}
where $\mu = \frac{1-\gamma}{\gamma} \alpha$. 
We can now rewrite the quality functions using only $p_{\rm in}$ and $\mu$.

\begin{restatable}{proposition}{TCutoffNew}\label{T1New}
For $r \in [0, T]$, the residual scoring functions in the profile and marginal case can be expressed, up to a constant, as 
\begin{subequations}\label{rho-functions}
\begin{align}
&\textstyle \rho^{\text{\sc prof}}(r) = \max(\log p_{\rm in}(r) - \log \mu, 0), \label{T1Q-prof}\\
&\textstyle \rho^{\text{\sc marg}}(r) = \smax(\log p_{\rm in}(r) -\log \mu, 0), \label{T1Q-marg}
\end{align}
\end{subequations}
where $\smax$ is the smooth maximum: $\smax(x,y) = \log(e^x + e^y)$. 
\end{restatable}
The proofs can be found in~\cref{A:proofs}. Note that we did not specify the likelihood for $r>T$. As we cannot adequately model the inlier-outlier likelihood ratio for this range, it is safer just to extend the expressions~\eqref{rho-functions} to an indefinite range $[0, \infty)$. Estimating $\theta$ by maximizing $Q(\theta)$ with these extended function becomes {\em M-estimation} (maximum-likelihood type). This resolves the gap between the true likelihood and M-estimators: the extended functions are, by design, the likelihoods in the relevant range $[0,T]$.

One immediate consequence is that the profile quality function can be given in the form of deciding inliers and scoring only them:
\begin{align}\label{Qprof-I}
\textstyle Q^{\text{\sc prof}}(\theta) = \sum_{i \in \I} (\log p_{\rm in}(r_i) - \log\mu),
\end{align}
which is a more explicit form and provides an insight that a decision about inlier is made internally early on.

Our next step is to go from the threshold $\mu$ on the density to the threshold $\tau$ on the residual value, which is possible when $p_{\rm in}$ is monotone.
\begin{lemma}\label{L:mon}
If $p_{\rm in}(r)$ is monotone non-increasing and $p_{\rm in}(0) > \alpha$ and $p_{\rm in}(T) < \alpha$ holds, there exists $\tau$ such that the set of inliers can be written as
$\I = \{i \mid r_i < \tau \}$.
If $p_{\rm in}(r)$ is also continuous, then $\tau$ is the unique root of $p_{\rm in}(\tau) = \frac{1-\gamma}{\gamma} \alpha$ and hence $\log \mu = \log p_{\rm in}(\tau)$.
\end{lemma}
Instead of solving equation $p_{\rm in}(\tau) = \frac{1-\gamma}{\gamma} \alpha$ for $\tau$ when given $\alpha$ and $\gamma$, we can consider $\tau$ as the primary independent parameter of the model and $\gamma$ and $\alpha$ as coupled, depending on $\tau$ and $p_{\rm in}$.
This view allows transparent comparisons of ML and M-type estimators and their related local optimization algorithms as it is fully sufficient for computing the scores as well as computing posterior probabilities of inliers.

\paragraph{Inlier Probabilities}
The posterior probability of inliers in the basic model expresses as:
\begin{align}\label{p_inliers}
\textstyle p(k{=}1 | r) =  \frac{p_{\rm in}(r) \gamma}{p_{\rm in}(r) \gamma + \alpha (1-\gamma)} = \frac{p_{\rm in}(r) }{p_{\rm in}(r) + \mu} = 
\frac{p_{\rm in}(r) }{p_{\rm in}(r) + p_{\rm in}(\tau)} = \sigm\Big(\log \frac{p_{\rm in}(r)}{p_{\rm in}(\tau)}\Big),
\end{align}
where $\sigm$ is the logistic sigmoid function.

Furthermore, when comparing which geometric model $\theta$ is better, the score function may be defined up to a positive scale and bias. 
We will therefore consider {\em normalized} versions that satisfy $\hat \rho(\infty) = 0$ and $\hat \rho(0) = 1$. The first connection we establish between robust and probabilistic methods is the following.

\begin{restatable}{proposition}{PRansac}\label{P:RANSAC}
For a uniform $p_{\rm in} = \beta$ with any $\beta > \alpha$ both the normalized profile likelihood quality function and the normalized marginal likelihood quality function match the {\bf RANSAC} quality function
$Q^\text{\sc ransac} = |\I|$.
\end{restatable}

\begin{figure}[t]
\centering
\includegraphics[width=0.5\linewidth]{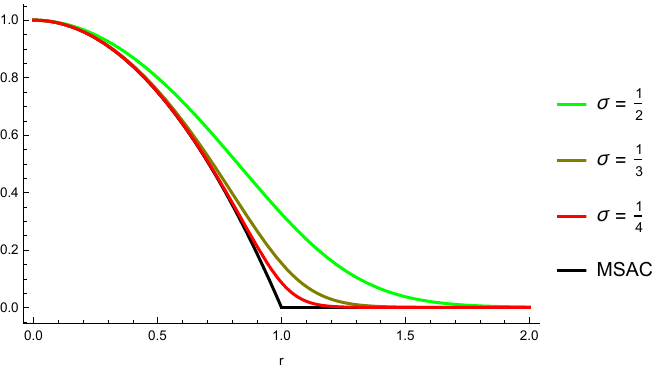}
\caption{\label{fig:msac-gu}
Comparison of profile score (equal to MSAC in this case) and marginal score with different scale for the Gaussian-Uniform mixture model. Both scores are shown for the decision threshold $\tau=1$.
}
\end{figure}

\subsubsection{Gaussian-Uniform (GaU) Mixture Model}\label{sec:Gauss-U}
Particularly simple expressions and neat algorithms are obtained for the basic model with the Gaussian inlier ray density, $\mathcal{N}(0, \sigma^2)$, which we expect to be of high relevance in practice due to normality tendency of all errors. 
\begin{restatable}{proposition}{PGaU}\label{P:GaU}
Under Gaussian inlier ray density and assuming conditions of \cref{L:mon} to hold we have the following:
\begin{itemize}
\item For any scale $\sigma$ (subject to the above conditions), the normalized profile scoring function matches {\bf MSAC}: $\hat \rho^{\text{\sc MSAC}}(r) = \max(1 - r^2 / \tau^2, 0)$.
\item The normalized marginal scoring function {\bf GaU} is: 
\begin{align}\label{GaU-rho}
\textstyle \hat \rho^{\text{\sc GaU}}(r) = \frac{1}{\beta}\smax\Big(\frac{\tau^2 -r^2}{2 \sigma^2 }, 0\Big), \ \ \ \ \text{where} \ \ \beta = \smax(\tau^2/2 \sigma^2, 0).
\end{align}
\item The posterior inlier probability is $p(k{=}1 \mid r) = \sigm\Big(\frac{\tau^2 - r^2}{2 \sigma^2 }\Big)$.
\end{itemize}
\end{restatable}
\noindent The Gaussian-Uniform mixture model is well known, however this presentation is novel in several aspects. 
MSAC was originally proposed as a robust M-estimator for Gaussian inliers, simply speaking, by truncating the Gaussian likelihood function~\cite{Torr-98}. we have now identified it with the profile likelihood of the mixture model.
Second we derived a simple "smoothed" version of the MSAC function which has not been used directly in RANSAC before. Thanks to parameterization with the decision threshold $\tau$, we can easily compare them for the same threshold as illustrated in \cref{fig:msac-gu}. The ratio $\frac{\sigma}{\tau}$ controlls the smoothing effect of $\smax$. 

\citet{torr2000mlesac} aslo marginalize the mixture model in $k$, but apply it in a different context. They consider the outlier density $\alpha$ and the inlier variance $\sigma^2$ to be fixed and estimate the inlier ratio $\gamma$ per image pair with EM for the purpose of scoring minimal samples. We propose the simplified expression $\rho^{\text{\sc GaU}}$ with fixed parameters $\tau, \sigma$ for both: scoring minimal samples and local optimization in $\theta$ considered next.

%



\subsection{Local Optimization}\label{sec:LO}
Once we found a good model $\theta$ by scoring minimal modles in RANSAC, it can be locally optimized to fit all the observations. A common practice is to decide the set of inliers and use non-robust estimation methods to fit to them or to optimize a robust criterion such as MSAC~\cite{torr2002bayesian} on all points. These approaches require setting the respective thresholds. However, once a quality function $Q(\theta) = \sum_{i}\rho(x_i, \theta)$ is defined for minimal samples, it is natural to continue to optimize the very same quality function. This becomes obscure in practice because {\em probabilistic modeling and robust statistics employ different parameterizations} and seemingly different optimization strategies.
Expectation maximization (EM) is a natural choice for estimating the parameters of a mixture model. 
IRLS is a widely used algorithm in robust regression and is applicable to a broad class of score functions \cite{maronna2019robust}. The IRLS iteration for maximizing $Q(\theta) = \sum_i \rho(r_i)$ is solving the problem
\begin{align}\label{IRLS}
\textstyle \theta^{t+1} = \arg\min_\theta \sum_i w^t_i r_i(\theta)^2,
\end{align}
where $w^t_i = -\frac{\rho'(r^t_i)}{r^t_i}$ are the weights using the model in the preceding iteration $r^t_i = r_i(\theta^t)$, \cite{maronna2019robust}. We discuss how this is derived in the proof of~\cref{P:EMIRLS} in~\cref{A:proofs}.

When optimizing the marginal likelihood in $\theta$ in the basic model, both EM and IRLS can be applied. We observe that they can be written in terms of $p_{\rm in}$ and $\tau$ only, exposing their similarities the most for a transparent comparison. 
\begin{restatable}{proposition}{PEMIRLS}\label{P:EMIRLS} An Iteration of EM (resp. IRLS) algorithm for optimizing marginal likelihood of the basic model in $\theta$ can be written as
\begin{subequations}\label{general-EM}
\begin{align}
\textstyle q_i^t = & \textstyle \sigm(\log p_{\rm in}(r_i(\theta^t)) - \log p_{\rm in}(\tau))\vphantom{\Big(}; \tag{Inlier probs} \\ 
\textstyle \theta^{t+1} = & \textstyle \argmax\limits_{\theta} \sum_i q^t_i \log p_{\rm in}(r_i(\theta)))\vphantom{\Big(} \tag{EM}\\
\text{resp.} \ \ \ \ \textstyle \theta^{t+1} = & \textstyle \argmin\limits_{\theta} \sum_i q^t_i \Big(\frac{d \log p_{\rm in}(\sqrt{R})}{d R}\Big|_{R = r_i(\theta^t)^2}\Big) r_i(\theta)^2. \tag{IRLS}
\end{align}
\end{subequations}
\end{restatable}
\noindent Both EM and IRLS are incomplete algorithms in that they do not specify what to do if the maximization in $\theta$ cannot be solved in closed form.
EM formulation has a weighted sum of $\log p_{\rm in}(r_i)$ in the objective, while IRLS further linearizes $\log p_{\rm in}$ as a function of $r_i^2$ in the current parameters and thus obtains a weighted sum of $r_i^2$ in its objective. However, the latter still results in a non-linear least squares minimization in $\theta$ in general\footnote{Because the residuals $r$ may depend on the model $\theta$ non-linearly.}. This particular approximation is reasonable when $\log p_{\rm in}(r)$ is approximately quadratic around $r = 0$, which connects to (locally) Gaussian errors.

\paragraph{IRLS-LMA} The advantage of IRLS formulation is that it casts the problem of finding $\theta^{t+1}$ as a weighted non-linear least squares, allowing the application of Gauss-Newton / Levenberg-Marquardt Algorithm (LMA).
In order to make IRLS~\eqref{IRLS} a complete algorithm, the vector of residuals $r(\theta)$ is linearized as $r_i(\theta) \approx r^t_i + J_i (\theta - \theta^{t})$, where $J_i = \frac{d r_i (\theta^t)}{d \theta}$. In this context, $\theta$ is assumed to be an unconstrained parameter in $\Real^m$. To represent \eg, an essential matrix, it is  necessary to parametrize it appropriately. Then the iteration must solve
\begin{align}
\theta^{t+1} = \arg\min_{\theta} \sum_{i=1}^{n} w_i (r^t_i + J_i (\theta - \theta^{t}))^2,
\end{align}
which is a weighted least squares problem with a closed form solution 
\begin{align}
\theta  = \theta^{t} -G^{-1} J^T\diag(w) r^t.
\end{align}
where $J$ is $n \times m$ matrix with rows $J_i$ and $G = J\T \diag(w) J$. The inverse of $G$ can be regularized in LMA as $(G + \lambda_1 \diag(G) + \lambda_2 I)^{-1}$, where $\lambda_1,\lambda_2>0$ are step-regularizing parameters, which are necessary when $G$ is not full rank (\ie, when a non-minimal parametrization $\theta$ is used) and can also be dynamically adjusted to ensure monotonic decrease of the objective in each iteration.

In the context of our basic mixture model, so-adopted LMA can be seen to use three approximations: lower-bounding the likelihood of the mixture model as in EM, approximating $\log p_{\rm in}(r)$ by $r^2$ and linearizing $r$ in $\theta$. If the inlier distribution $p_{\rm in}(r)$ is Gaussian, the middle approximation is exact and the method simplifies.

%

\begin{restatable}{proposition}{PGaUEM}\label{P:GaU-EM}
For the GaU marginal likelihood, EM and IRLS schemes coincide, the update step of $\theta$ is given by 
\begin{subequations}\label{GaU-EM}
\begin{align}
\textstyle q^t_i = p(k_i{=}1 \mid r(x_i, \theta^t)), \label{q_i^t}\\
\textstyle \theta^{t+1} = \argmin_{\theta} \sum_{i=1}^{n} q^t_i r(x_i, \theta)^2.
\end{align}
\end{subequations}
\end{restatable}
%
Thus, if we use GaU marginal likelihood for scoring minimal hypothesis, it is straightforward to use it as well for local optimization in $\theta$ for refining the hypothesis inside the RANSAC loop or for the final result. For the purpose of understanding the IRLS scheme of MAGSAC++, note that the weights \eqref{q_i^t} here are the posteriori probabilities {\em of points being inliers} for the current model.

\section{Analysis of MAGSAC/MAGSAC++}\label{sec:M++}
In the core of MAGSAC(++) methods \cite{barath2019magsac,barath2020magsac++,barath2021marginalizing} are non-conventional robust scores/estimators for geometric fitting problems, which stand out from the above probabilistic framework. Their remarkable benchmark performance~\cite{ivashechkin2021vsac} motivates studying their design principles.
The analysis below explains why these design principles are not sound and can give the same result as the GaU model at best, when the derivation errors mutually cancel, and would lead to poor estimators otherwise.


\subsection{The Likelihoods}\label{sec:MAGSAC-likelihood}
The starting point in~\cite{barath2021marginalizing} is observing that, typically, the residuals of the inliers are calculated as the Euclidean distance from model $\theta$ in some $\nu$-dimensional space (\eg, the re-projection error). The errors along all axis of the space are assumed to be independent, normally distributed, implying ({\red attention, ERROR follows!}) that the residuals are $\chi_\nu$ distributed.
This informal reasoning misses the difference between the point-to-point distance and point-to-model distance. The distance to a manifold defined by $d_g$ equations is $\chi_{d_g}$-distributed instead of full $\chi_\nu$, becasue the true correspondance satisfying the model $g(\bar x; \theta) = 0$ is unknown and has to be maximized or marginalized out. See \cref{A:Line2D} in \cref{A:MAGSAC++} for the maximization approach in the elementary 2D line fitting problem of~\cite{barath2021marginalizing}. This leads~\cite{barath2021marginalizing} to considering $\chi_4$ distributed inlier residuals for both the relative pose problem (Essential / Fundamental matrix) and Homography, because in both cases observations are pairs of points in two views and thus $\nu=4$. The correct distributions are $\chi_1$ and $\chi_2$, resp. (see~\cref{fig:inliers}).

Suppose nevertheless, that inlier residuals are indeed $\chi_\nu$ distributed. This is the case for the absolute pose problem, where the true projection $\bar x \in \Real^2$ is known exactly, observations $x \in \Real^2$, so $\nu=2$, and assuming $x \sim \N(\bar x, \sigma^2 I)$, it follows that $\|x - \bar x\|/\sigma  \sim \chi_2$. Another example is that of registering two point clouds in $\nu=3$ dimensions.
Since the distribution of residuals is from a known family, \citet{barath2019magsac,barath2021marginalizing} propose ({\red attention, ERROR follows!}) to consider the likelihood of inlier residuals according to this distribution as the basis for their robust scoring.
The issue is that {\em the likelihood of residuals according to the distribution of residuals does not lead to a consistent estimator}. This is because residuals depend on the model parameters to be estimated. An immediate consequience is that the likelihood of the ground truth model with zero residuals would be zero (becasue the density of $\chi_\nu$ = 0 for $\nu\geq 2$). A detailed mathematical explanation why this happens is given in \cref{example-ML-transform} of \cref{A:MAGSAC++}.

MAGSAC~\cite{barath2019magsac} defines the likelihood of chi-distributed residuals. Unlike observations $x$, which remain fixed, the residuals change together with $\theta$.
They use it in the mixture model with uniform outliers~\cite[Eq. 4]{barath2019magsac} and integrate the resulting score over unknown $\sigma$. The above concerns apply directly in this case, including the pathology that the likelihood of the ground truth model to have zero residuals is zero.
The estimation results would be yet different if we consider, for instance, the likelihood according to the distribution of squared residuals.

MAGSAC++ uses the problematic likelihood of residuals in an indirect way. As we will show below, by stacking more unsound steps, it obtains an estimator practically unrelated to the initially assumed model of inlier residuals and is thus unaffected by the pitfall. 
%
%
However, the lesson that we should take away is that it is a bad idea to consider the likelihood of model-dependent residuals as a starting point in either case.


\subsection{MAGSAC++ Estimator}

In the reminder of the section we give a concise exposition of MAGSAC++ derivation, highlight the errors and analyze their net effect.

\subsubsection{Derivation Step 1: Marginalization}
Let $p_{\chi}$ be the density of chi distribution with $\nu$ degrees of freedom and let $p_{\bar \chi}(x)$ be the truncated chi density with support $[0, \kappa]$. 
The inlier density is introduced as a scaled truncated chi density:
$g(r \; | \; \sigma) = \frac{1}{\sigma}p_{\bar \chi}\big(\frac{r}{\sigma}\big)$.
The noise scale $\sigma$ is considered to be unknown and uniformly distributed in $[0, \bar\sigma]$, where $\bar \sigma$ is the maximum value (denoted $\sigma_{\rm max}$ in~\cite{barath2021marginalizing}). Integrating over $\sigma$ leads to the scale-marginalized inlier density
\begin{align}\label{magsac-w}
p_{\rm in}(r; \bar\sigma) = \frac{1}{\bar \sigma}\int_{0}^{\bar\sigma} g(r \; | \; \sigma) d \sigma.
\end{align}
We note that $\bar\sigma$ is effectively the scale parameter of this new distribution because there holds scale identity $p_{\rm in}(r; \bar\sigma) = \frac{1}{\bar \sigma}p_{\rm in}(r; 1)$ (\cref{A:MAGSAC++}:P1).
The marginalization idea thus reduces to using $p_{\rm in}(r; \bar\sigma)$ with scale $\bar\sigma$ instead of $g(r\;|\; \sigma)$ with scale $\sigma$.
Unlike the chi density, this scale-marginalized $p_{\rm in}$ is monotone decreasing. It is visualized in~\cref{fig:magsac} (left) for different values of $\nu$.
The marginalization step effectively masks the issue with the chi likelihood being zero at $0$, 
except for the case $\nu=1$, where $p_{\rm in}(r; \bar\sigma)$ is unbounded at zero.

\subsubsection{Derivation Step 2: Probability of Inliers and IRLS Weights}
In the next step~\citet{barath2021marginalizing} define the ``likelihood of a point $x_i$ being an inlier'' ({\red attention, ERROR follows!}) equal to  $g(r(x_i; \theta) \; | \; \sigma)$ (this is seen by comparing $P(p\mid \theta_\sigma, \sigma)$ with $g(r \mid \sigma)$ in \cite{barath2021marginalizing}).
It is unclear whether the interpretation ``likelihood of a point $x_i$ being an inlier'' or the equation itself is incorrect, but combined together this leads to absurd conclusions. The boolean expression "point $x_i$ is an inlier" is denoted here as $k_i=1$ and its likelihood (or probability) given the value of the residual is $p(k_i{=}1 \mid r_i; \theta)$. The statement implies $p(k_i{=}1 \mid r_i; \theta) = p_{\rm in}(r_i; \theta)$, equating probability of an event with a density over a continuous domain. A density is not limited to be less than $1$ and is even unbounded for $\chi_1$ when marginalized over $\sigma$.

Next, IRLS weights are designed. IRLS is an optimization scheme like Gauss-Newton. Normally, it is applied to a given quality function such as the likelihood or a robust $\rho$ function \cite{maronna2019robust} and the weights are derived from that function as we did in~\cref{P:EMIRLS,P:GaU-EM}. \citet{barath2021marginalizing} use it the other way around: they propose the weights directly, effectively setting them equal to the (scale-marginalized, truncated) density: $\tilde w_i^t = p_{\rm in}(r(x_i, \theta^t); \bar \sigma)$.
While the posterior probabilities of inliers commonly occur as weights in EM and IRLS (see~\cref{P:EMIRLS,P:GaU-EM}), using the density as weights is completely arbitrary. It is not more justified than using \eg a squared density. And the same weights cannot be both the density and the posterior inlier probability! Assuming merely a proportionality between the two (which would be enough, since the scale of weights in IRLS does not matter) has strong implications. It can be shown (\cref{A:MAGSAC++}:P2) that in this case the outlier density must necessarily be of the form $p_{\rm out}(r) = a - b p_{\rm in}(r)$ for some $a,b>0$, \ie conveniently proportional but opposite in sign to the inlier density and in high contrast to the claim that MAGSAC++ is free from assumptions about the outlier distribution~\cite{barath2021marginalizing}. 

\subsubsection{Derivation Step 3: Score from IRLS}

The weight and the scoring functions in IRLS are related by the differential equation $\frac{-\rho'(r)}{r} = \tilde w(r)$. 
Integrating it results in the MAGSAC++ score 
\begin{align}
\rho(r; \bar \sigma) = -\int_{0}^{r} x p_{\rm in}(x; \bar \sigma) d x.
\end{align}
See \cref{A:MAGSAC++}:P3 for a closed-form expression through Gamma functions.
Because this score is derived from the $\chi_\nu$ likelihood distribution it is implied that ({\red attention, ERROR follows!}) it is a robust estimator suitable for $\nu$-dimensional geometric problems.
This is not true: because $\chi_\nu$ distribution was plugged into IRLS scheme in an arbitrary way, the connection between the resulting score and the initial problem and its dimensionality $\nu$ is arbitrary. The question is: to which likelihood does this score truly correspond? We answer it next.

\subsubsection{Alignment with GaU Model}
\begin{figure}
\centering
\setlength{\tabcolsep}{5pt}
\begin{tabular}{cc}
\scriptsize Base weight function $\tilde w(x) = p_{\rm in}(x; 1)$ & \scriptsize Base Score function $\rho(x; 1)$\\
\includegraphics[width=0.46\linewidth]{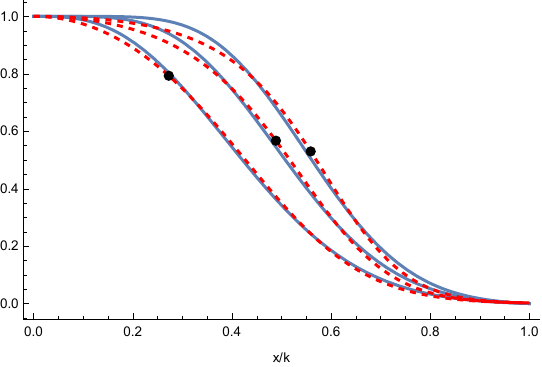}&
\includegraphics[width=0.46\linewidth]{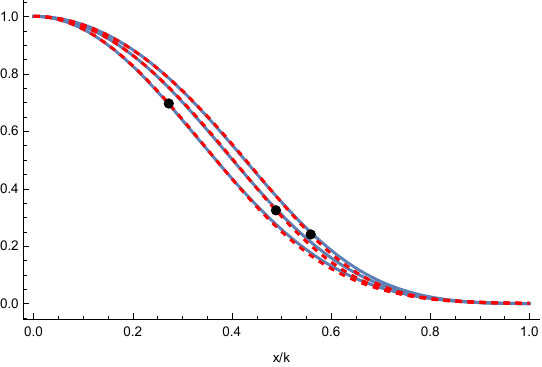}
\end{tabular}
\caption{\label{fig:magsac}
{\em Left:} Normalized weight of MAGSAC++ (blue) for $\nu=4,6,8$ and $\bar \sigma = 1$ and the fitted inlier posterior probability function of Gaussian-Uniform mixture model (dashed red). We fit both $\tau$ and $ \sigma$ of GaU model.
{\em Right:} Corresponding normalized score functions. The respective 99'th quantile $\kappa = 3.64, 4.1, 4.45$ and $\tau = \bar \sigma \kappa = \kappa$ for MAGSAC++. The fitted parameters $(\tau,\sigma)$ of GaU are, respectively, $(1, 0.96), (2, 1), (2.51, 1.06)$. Fitted threshold parameters $\tau$ of GaU model are visualized by the black dots in this normalized plot.
}
\end{figure}

It turns out that the ad hoc weights $\tilde w(r) = p_{\rm in}(r; \bar \sigma)$ can be rather accurately approximated by the posterior inlier probabilities of the GaU model as shown in~\cref{fig:magsac} for several common values of $\nu$. In particular, we make the following formal statement.
\begin{proposition}\label{P:MAGSAC-GaU}
For $\nu=4$, the family of MAGSAC scores $\rho(r; \tau)$ parameterized by $\tau$ with $\bar \sigma = \tau/\kappa$ where $\kappa$ is the 99'th quantile $\chi_\nu$, is numerically equivalent to the family of GaU normalized marginals scores $\hat \rho^{\rm GaU}(r; \tau, \sigma)$~\eqref{GaU-rho} parameterized by $\tau$ with $\sigma = 0.96 \tau \approx \tau$.
\end{proposition}
\begin{proof}
Thanks to the scale identities we have proven (\cref{A:MAGSAC++}:P3), it is sufficient to demonstrate the approximation for a single scale $\bar \sigma = 1$. For this scale we fit the weight function of MAGSAC++ with the inlier posterior of GaU model by optimizing over $\tau$ and $\sigma$. The fit is very accurate for $\nu=4$, as shown in \cref{fig:magsac} (left).
With this approximation, the heuristic IRLS scheme becomes the IRLS scheme of the GaU model and the resulting scoring function $\rho$ almost perfectly aligns with the marginal score of the GaU model \cref{fig:magsac} (right). We thus see a numerical equivalence, implying that the two methods must be indistinguishable in practice.
\end{proof}

%

\subsubsection{MAGSAC++ Explanation and Conclusions}\label{sec:MAGSAC-conclusion}
We have identified a number of errors in the derivation of MAGSAC++ and have arrived at the following explanation of how these errors lead to reasonable results:
\begin{enumerate}[noitemsep]
\item It starts with the assumption of $\chi_\nu$-distributed inliers, which miscounts degrees of freedom and does not lead to a sound likelihood. 
\item Luckily, when the above density of inliers is marginalized over the uniformly distributed noise scale $\sigma$, it becomes rather similar to the sigmoid-shaped posterior of a GaU model (\cref{fig:magsac}, left).
\item Once scale-marginalized density of inliers is similar to the inlier posterior of a GaU model, when plugged into the IRLS scheme as weights, it recovers the correct IRLS scheme of the GaU model~\eqref{GaU-EM}, and hence the score recovers the marginal scoring function of the GaU model (\cref{fig:magsac}, right).
\end{enumerate}

In the cases when MAGSAC++ was reported to perform well (relative pose and homography estimation), the degrees of freedom $\nu$ are set to $4$ and the GaU approximation above holds accurately, implying that it is numerically equivalent to the likelihood of a well-known GaU model, to which we gave a simple threshold-parameterized expression. 
It is worth emphasising that it is a lucky coincidence. There is no sound design principle behind MAGSAC++ that can be applied to other problems or even to the same problems with different assumptions of the scale distribution. If any of the ingredients are changed, nothing good may be expected. For example:
\begin{itemize}
\item Start from a Gaussian distribution of residuals (as appropriate for the 2D line model or epipolar geometry, \ie fixing the first error). We will get an infinite weight at 0.
\item Start from a $\chi_\nu$ distribution but assume that the scale is known (or concentrated around a known value). We will get a zero weight at 0. IRLS scheme with nonmonotone weights $\tilde w = g(r | \sigma)$ would likely behave poorly and correspond to a nonsense score function.
\item Do not truncate the distribution, \ie set the truncation quantile $\kappa$ to a very large value, and $\tau=\sigma \kappa$ for an outlier-free scenario. The estimator stays ``robust'', incorrectly suppressing a part of the inliers.
\end{itemize}
%
%
Contrary to its name, MAGSAC is {\em not scale-marginalizing} (at least, not in any relevant sense), because in the end it is numerically equivalent to GaU model, which has Gaussian inlier residuals (and not scale-marginalized Gaussian). It has a dependence on the degrees of freedom $\nu$ and the maximum noise scale $\bar \sigma$, but {\em without a useful meaning}. \citet{barath2021marginalizing} set the scale $\bar\sigma$ as $\bar\sigma = \threshold / \kappa$ adopting $\tau$ as the primary parameter.
This threshold has the interpretation of a fixed quantile of the inlier distribution. In the normalized scale plot in \cref{fig:magsac} this is the point at $x/\kappa=1$, where the weight and score functions vanish. It {\em does not} have an interpretation of a decision threshold for determining inliers and is not suitable as such.


\section{Evaluation Methodology}
Conventionally, RANSAC is executed sequentially, drawing a minimal sample, solving for the model hypothesis, and keeping track of so-far-the-best model according to the score. It can stop before the iteration limit if a confidence criterion is satisfied. RANSAC is further enhanced with local optimization (locally improving each candidate hypothesis by various techniques, \eg, \cite{chum2003locally,lebeda2012fixing}) and by the final optimization of the result. The stopping criterion, local and final optimization techniques need to rely on some notion of inlier-outlier threshold for using non-robust fitting or on a scoring function. However, often a different scoring function is used for refining from the one used in the selection loop. For example, MAGSAC++ employs MSAC score for refining, which, among other things, requires a different threshold parameter.

When the complete pipelines are compared~\cite{barath2021marginalizing,barath2022learning,Jin-21}, the advantage of MAGSAC++ or, say, the MQNet~\cite{barath2022learning} pipeline, might be due to other factors than the score functions they propose.
We created a basic plain pipeline in PyTorch~\cite{paszke2017automatic} which targets comparison of scoring functions and takes advantage of parallelism and convenience of Python.
Instead of a termination criterion, we generate a pool of minimal geometric models.
We evaluate all scoring functions on all geometric models. This allows for a direct comparison of scoring functions and speeds up computations since the residuals need to be computed only once for each candidate model.

\paragraph{Cross-Validation}
The scoring functions we consider have one free threshold hyperparameter. It turns out that choosing this threshold is a cornerstone of an accurate comparison. 
Historically, there were no large validation sets available and the common colloquial view was that RANSAC should work for the user ``without a validation set'' assuming that good threshold values for a method are known. Similarly, it was expected to work reliably after choosing the threshold using a small validation set. It is a common practice to check only few thresholds and compare the methods using only a single validation-test split. Unfortunately, such a comparison may indicate that one method is better than another depending on which particular validation set was used and which thresholds were tried. The later is also important, considering that different methods interpreter the "threshold" parameter differently.

There are two ways to make the evaluation more objective: either consider a large enough validation set, drawn from the same distribution as the test set, or consider multiple random validation sets and compare the performance of methods in expectation. We will conduct experiments in both settings while also densely sampling threshold values.


\subsection{Large Validation Set}
In this setup we assume that a large validation set is available, obtained in the similar setup as the test set (\ie, same kind of features, filtering of correspondences, \etc.). In such a case, for each method we can pick the validation-optimal threshold hyperparameter with a high precision. 
This setup may not be feasible in the end-user applications, but it allows us to objectively compare methods from the perspective of their performance with an ``optimally tuned'' threshold. It is also relevant in the context of comparison with  deep learning methods such as MQNet~\cite{barath2022learning}, which use a very large training set.
Because the scale of threshold is method-dependent, it is not sufficient to check a limited grid of thresholds such as 12 selected values in the range $[0, 20]$ in~\cite{Jin-21}. Towards this end we use a grid of 200 threshold values. This might be computationally infeasible for existing pipelines, but we take advantage of parallelism and histogram representation of additive scoring, allowing us to evaluate all thresholds while computing residuals only once.

For scoring methods like MSAC and GAU, the scoring function $\rho$ has a simple analytic expression.
However, once the parameters are chosen, the complete function $\rho$ can be also precomputed as a look-up table. We can discretize residuals in the range $[0, \tau_{\rm max}]$ for a suitable larger $\tau_{\rm max}$ (\eg 10 px). Let the interval $[0, \tau_{\rm max}]$ be partitioned into $K$ uniform bins $b_k$ with centers $\mu_k$. Let $w_k = \rho(\mu_k)$ be discretized representation of $\rho$. We can then express additive quality function of residuals using the histogram of residuals
\begin{align}
\textstyle Q(\theta) = \sum_{i} \rho( r_i ) = \sum_{i} \sum_{k} w_k \bbleft r_i \in b_k \bbright = 
\sum_{k} w_k \sum_{i} \bbleft r_i \in b_k \bbright = w\T h,
\end{align}
where $h$ is the count of residuals in each bin: $h_k = \sum_{i} \bbleft r_i \in b_k \bbright$. 

If the function $\rho$ is simple to compute, the histogram representation does not lead to a speed-up of evaluating a single quality function as it still needs to process all correspondences. It pays off however when evaluating multiple scoring functions and especially for multiple thresholds. For $T=200$ thresholds and $K=500$ bins, for each method we form a matrix $W$ of size $T \times K$ and compute the scoring results for all thresholds as the product $W h$. Computing residuals, histogramming and scoring for a pool of minimal models can be done in parallel on GPU.

\subsection{Small Random Validation Set}\label{sec:small-method}
Let us assume that a potential user of RANSAC has only a few validation samples available to adjust the thresholds. 
In this scenario method A may appear better than method B in tests only by chance, due to a particular choice of the validation set. To recommend method A over method B to users in general, we need to look at the {\em test performance in the expectation over randomly drawn validation set}.

It is possible to compute such expected test performance in an efficient manner as follows.
For each image pair we can precompute the pose error of every method for every possible threshold value (\eg, 200 thresholds). The pre-computing procedure draws 1K minimal models per image pair, scores all of them using all thresholds, selects the best-scoring model for each threshold and evaluates the model-GT error (pose error) of that best model.

Using these precomputed errors, we proceed as follows. We draw a subset $V$ of image pairs from two random scenes for validation and consider all other scenes for test. On the validation set we compute median pose error using the precomputed error values for all thresholds, and select the best threshold $\tau(V)$.
Then we compute median pose error on the test set, $e(\tau(V))$, for the selected threshold, again using precomputed values. This can be repeated 1000 times to estimate the expected test error $\mathbb{E}[e(\tau(V))]$ in under a minute.

\subsection{Threshold Sensitivity}
Even when the test performance is the same, a method that is less sensitive to the choice of the threshold may be preferred in practice. An important question is how to quantify this sensitivity.

Prior work~\cite{barath2021marginalizing} assessed the sensitivity to the threshold based on the plots analogous to the validation error versus threshold as in \cref{fig:xval}. Namely, they compare the area under the graph of mean average accuracy (mAA) versus threshold. However, the sharpness of this graph and the respective area may differ depending on how the hyperparameter is actually used to compute the score --- which linear or non-linear transformation it undergoes. As a trivial example, we may redefine MSAC as $\max(0, 1-r^2/(\tau^2/3))$ and it will be much more insensitive to the choice of the threshold according to~\citet{barath2021marginalizing}.

We propose the following methodology, which is more directly interpretable and is invariant to any monotone remapping of the hyperparameter. 
We propose to measure sensitivity by the variance of the test performance, $\mathbb{V}[e(\tau(V))]$, which can be estimated alongside $\mathbb{E}[e(\tau(V))]$ discussed above. We propose that this variance captures the right notion of sensitivity: how much the test error depends on the sub-optimal choice of hyperparameter due to the limited validation data. This is exactly the sensitivity a user with a small validation set would care about.

\section{Experiments}\label{sec:experiments}
We have conducted multiple evaluations on homography and relative pose estimation problems.
Our implementation is available at \url{https://github.com/shekhovt/RANSAC-Scoring}. Secondary details of the experiments can be found in \cref{A:exp_detail}.

\begin{figure}[t]
\centering
\begin{tabular}{cc}
\includegraphics[width=0.49\linewidth]{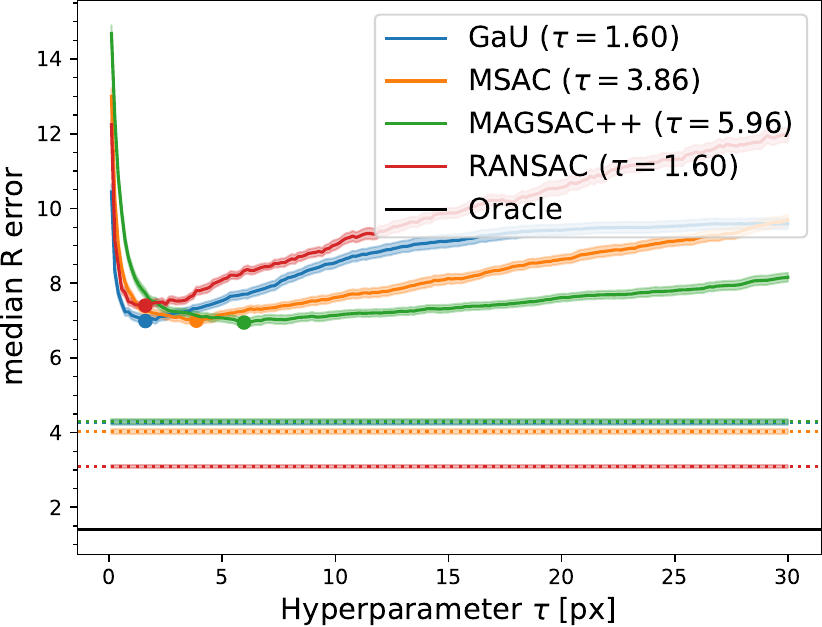}&
\includegraphics[width=0.49\linewidth]{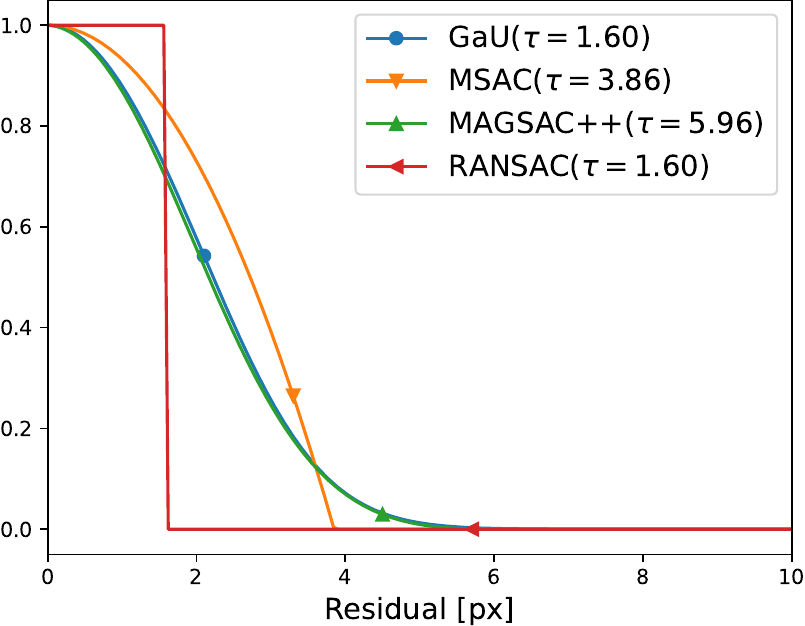}
\end{tabular}
\caption{
Validation on the homography estimation benchmark (HEB).
{\em Left:} validation plot. Dotted horizontal lines show score-based oracles: selecting the best threshold per image pair. 
Shaded areas show $95\%$ confidence intervals with respect to the validation set (here and below by BCA bootstrap).
{\em Right:} selected scoring functions.
\label{fig:validation_HEB}}
\centering
\begin{tabular}{cc}
\includegraphics[width=0.49\linewidth]{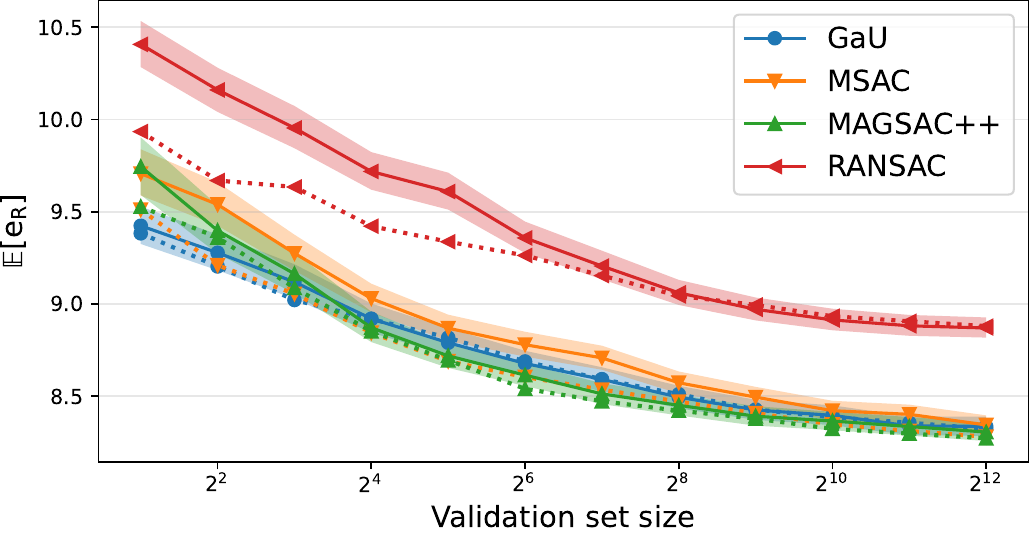}&
\includegraphics[width=0.49\linewidth]{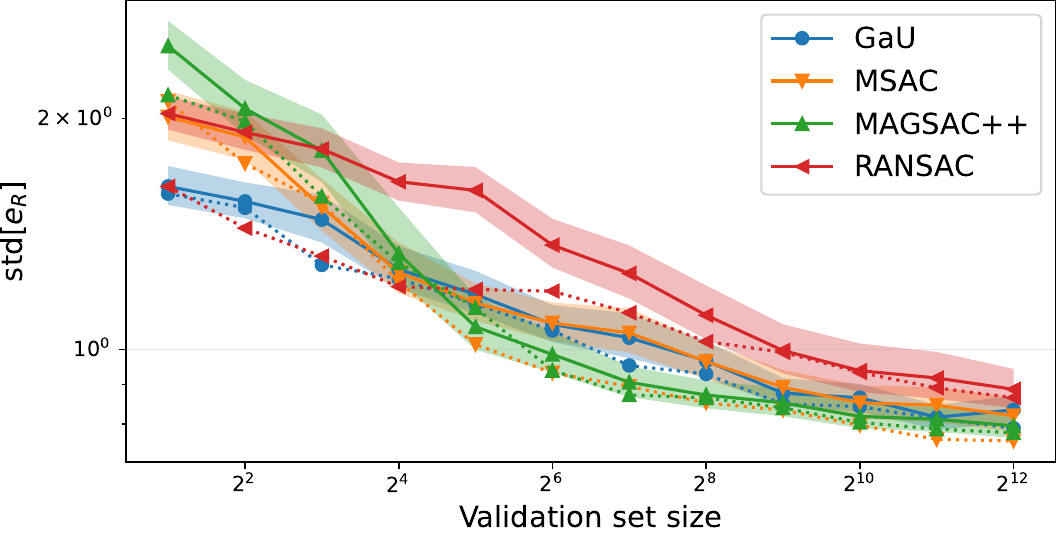}
\end{tabular}
\caption{
Expected test rotation error ({\em left}) and its std ({\em right}) for homography estimation on HEB. Confidence intervals are w.r.t. cross-validation trials.
\label{fig:expectd_HEB}}
\end{figure}
\subsection{Homography Estimation}\label{sec:HEB}
First, we present the evaluation on the large-scale Homography Estimation Benchmark (HEB)~\cite{Barath_2023_CVPR}.
There are total of 10 scenes with precomputed correspondences and ground-truth relative poses.
For tractability, we have randomly sampled 2k image pairs from each scene and generated 1k valid minimal models per pair.
For each image pair, scoring method and a 200 thresholds in $[0.1, 30]$, we have precomputed the rotation error of the best homography. Details of the model sampling and metrics are provided in \cref{A:exp-metrics,A:exp-HEB}. MAGSAC++ is evaluated using $\nu=4$, as in the authors implementation, and GaU model uses $\sigma=\tau$ motivated by~\cref{P:MAGSAC-GaU}.

\paragraph{Large Validation Set}
For the large validation set experiment, we have used 5 randomly selected scenes resulting in 10k image pairs. \cref{fig:validation_HEB} shows the median rotation error (in degrees) versus thresholds for different scoring functions and the optimal scoring functions $\rho(r)$ corresponding to the chosen thresholds. It is seen that while the optimal thresholds differ per methods, the best achievable performances are similar, with only RANSAC being inferior. The optimal scoring functions of MAGSAC++ and GaU coincide, as expected from the analysis in \cref{sec:MAGSAC-conclusion}.
The plot also includes the {\bf Oracle} method, which uses the true pose error as the scoring function, and score-based oracles, which select the best threshold per image pair. They allow to judge the quality of minimal models in the pool and give lower bounds on the achievable performance. The test performance will be evaluated in the small validation set experiment below and also, in more detail, for the relative pose problem in \cref{sec:PhotoTourism}.


\paragraph{Small Validation Set}
In this setting we perform the following cross-validation protocol. We draw 2 scenes as validation at random, sample $n$ image pairs from them to estimate the optimal threshold for each method, and evaluate the test error on the remaining 8 scenes. This is repeated for 1000 trials and for different sizes of the validation set $n$. \cref{fig:expectd_HEB} shows the expected (on average over trials) test rotation error and its standard deviation versus $n$. 

With a few validation pairs (the plot starts with 2), the estimation procedure is sensitive to the range in which the threshold is searched. As we know from the analysis, MAGSAC++, by design, uses a factor of $3.6$ greater threshold than GaU. The lower limit of $0.1$ contributes more prior information to GaU and the upper limit of 30 contributes more prior information to MAGSAC++. For a more fair comparison, we adjusted the set of candidate thresholds: for MAGSAC++ the search range is reduced to $[0.3, 30]$ and for all other methods the search range is reduced to $[0.1, 10]$. This compensates for differences in scaling. The results for the adjusted range are shown in dotted lines in \cref{fig:expectd_HEB}. The results of all methods improve, as expected, because tightening the search interval makes the threshold more reasonable for validation cases where the best threshold is found on the boundary of the search interval. It is seen that the choice of the interval has a significant impact on the performance for small $n$.

On the basis of the above experiment, we can draw the following conclusions. Vanilla RANSAC performs substantially worse than other scores. The performance of other methods, after the search range adjustment, is similar, with MAGSAC++ and GaU being equivalent, as expected. MSAC is slightly worse for small $n$ but the difference vanishes quickly with more validation pairs. Sensitivity analysis in \cref{fig:expectd_HEB} (right) shows that none of the methods is more sensitive than the other.


\subsection{Relative Pose Estimation}\label{sec:PhotoTourism}

This section presents experiments for the relative pose (essential matrix) estimation, extending the homogaphy study in several ways: we will study test set performance in dependnece of the number of samples and will include local optimization methods.

\paragraph{Dataset}
The experiments are conducted on the PhotoTourism dataset\footnote{CVPR IMW 2020 PhotoTourism challenge \cite{snavely2006photo}.} created by COLMAP, a globally optimized structure-from-motion 3D reconstruction from internet images~\cite{schonberger2016structure}. It contains large outdoor sights observed by multiple user cameras.
We use the ground-truth poses provided by~\citet{Jin-21}.
We consider correspondences with RootSIFT features matched with mutual nearest neighbors\footnote{Provided by \textit{RANSAC in 2020 Tutorial}~\cite{tutorialdata}.} and deep learning correspondences obtained with SuperPoint (SP)~\cite{detone2018superpoint} and SuperGlue (SG)~\cite{sarlin2020superglue}.  

\paragraph{Scoring Functions}
In addition to vanilla RANSAC, MSAC, MAGSAC++ and GaU score functions, we evaluate also a learned score function. The learning is performed by maximizing the likelihood ({\bf ML}) of correspondences under the inlier-outlier mixture for the true geometric model $\theta$, detailed in \cref{sec:learned}. 
As before, we also include the {\bf Oracle} method that uses the GT pose error $e$ as the criterion. 

\paragraph{}
Further details of the experimental setup and performance metrics are given in~\cref{sec:A-PhotoTourism-details}. %

\subsubsection{Validation and Threshold Sensitivity}\label{PhotoTourism-val}
\begin{figure}
\centering
\setlength{\tabcolsep}{0pt}
\begin{tabular}{ccc}
\multicolumn{2}{c}{\small PhotoTourism RootSIFT}\\
\scriptsize Validation  & \scriptsize Selected Scoring Functions \\
\begin{tabular}{c}\includegraphics[width=0.49\linewidth]{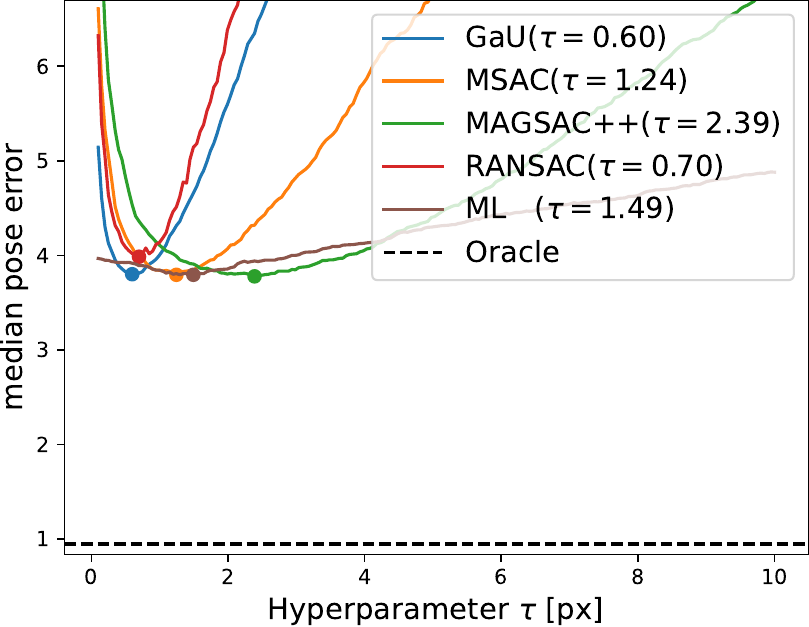}\end{tabular}
		& \ \
\begin{tabular}{c}\includegraphics[width=0.49\linewidth]{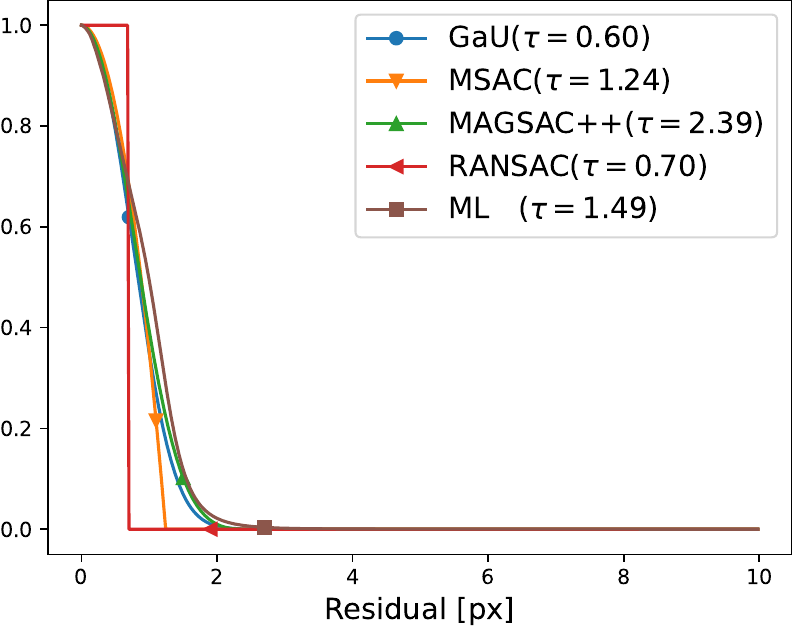}%
\end{tabular}
\end{tabular}
\caption{Validation of the threshold hyperparameter. {\em Left}: median error versus threshold.
{\em Right:} Scoring functions corresponding to selected thresholds. 
\label{fig:xval}
}
\centering
\begin{tabular}{cc}
\includegraphics[width=0.49\linewidth]{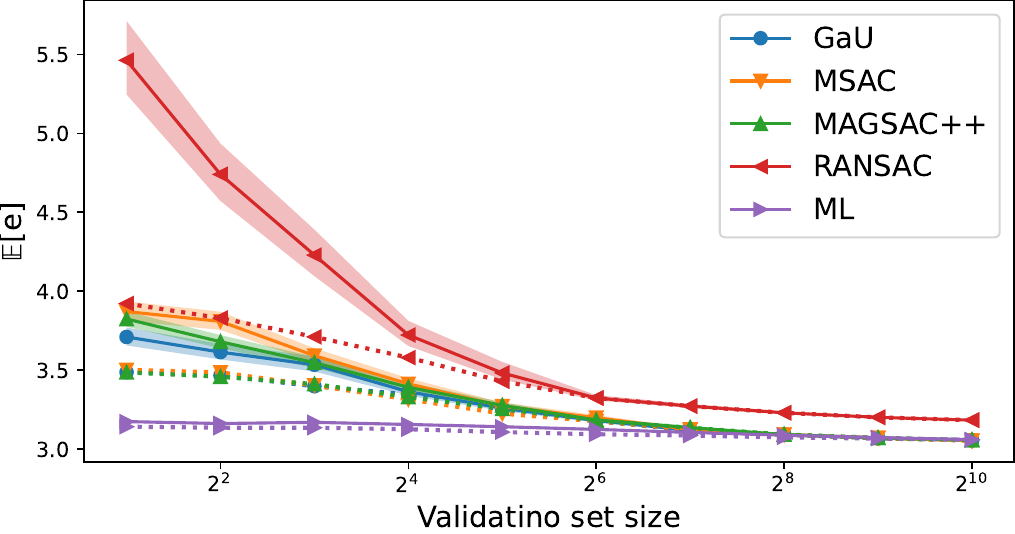}&\ \
\includegraphics[width=0.49\linewidth]{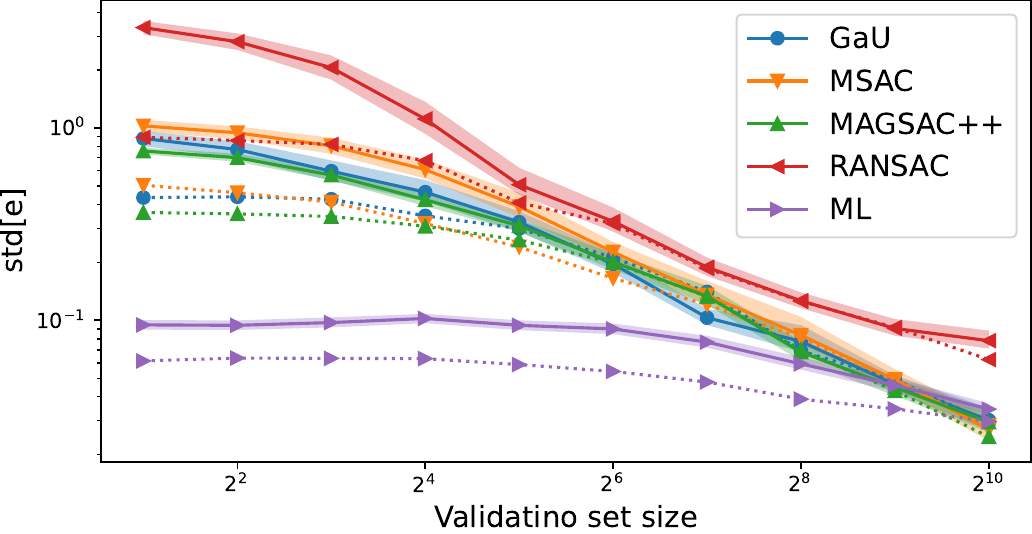}
\end{tabular}
\caption{
Expected test pose error ({\em left}) and its std ({\em right}) for PhotoTourism RootSIFT.
\label{fig:expectd_RootSIFT}}
\end{figure}

We first perform dense threshold validation using 11000 validation pairs of PhotoTourism (1000 per scene). 
\cref{fig:xval} shows the validation plots obtained. They confirm the same findings as for homography estimation in \cref{fig:validation_HEB}: methods except RANSAC, including the maximum-likelihood estimated scoring function, achieve the same performance and find very similar scoring functions $\rho(r)$. RANSAC is again substantially worse. Similar plots are observed for SPSG features in \cref{fig:validation_SPSG}, additional experiments using ETH3D and LAMAR datasets (\cref{A:ETH3D_LAMAR}) and for fundamental matrix estimation on KITTI (\cref{fig:KITTI-val}). 

We also performed small random validation set experiments similar to \cref{fig:expectd_HEB} for homography estimation. The results for PhotoTourism RootSIFT are shown in \cref{fig:expectd_RootSIFT} and for SPSG in~\cref{fig:expectd_SPSG}. They confirm the same conclusions as for homography estimation: RANSAC is substantially worse in expectation, while other methods perform similarly after adjusting the search interval.
%
Sensitivity analysis in \cref{fig:expectd_RootSIFT} (right) also shows quite similar results.
RANSAC with an unadjusted search range has a significantly higher sensitivity, but all other methods have similar sensitivity, especially after adjustment. 
The ML method is an exception: it has used the training scene of PhotoTourism to fit the shape of the scoring function, therefore comparing to it in \cref{fig:expectd_RootSIFT} is not objective. Nevertheless, it is satisfactory that ML learning achieves the same quality as the discriminative selection and is robust to the choice of its threshold hyperparameter, which we think is due to a heavier tail learned distribution~\cref{fig:learned}.

\subsubsection{Test Set Evaluation}\label{sec:PhotoTourism-test}
We will now evaluate the scoring functions shown in~\cref{fig:xval} with the validated thresholds on the test set of PhotoTourism. We will address the following questions: 1) how do the methods compare in selecting the best model from minimal samples on the test set, 2) how the performance depends on the number of samples drawn, and 3) how do local optimization methods using different scoring functions compare.

\paragraph{Local Optimization Methods}
We implemented IRLS with the Levenberg-Marquard Algorithm (IRLS-LMA, \cref{sec:LO}) for GaU score using a parameterization of the essential matrix with $(R,t)$, where the rotation is parameterized with quaternion.
We use automatic differentiation in PyTorch to compute the necessary Jacobians.
We compare our implementation to SOTA implementation of local optimization from PoseLib~\cite{PoseLib}: ``Truncated'', which is an IRLS scheme for MSAC score and ``Lee-Zach'' which is a recent SOTA scheme~\cite{Le-Zach} for the same MSAC score\footnote{Other loss functions implemented in PoseLib: Huber and Cauchy performed substantially worse when used with the same scale.}. We used these methods with our validated threshold $\tau$ for MSAC. 
All methods run for a maximum of 25 iterations\footnote{A larger number of iterations does not noticeably improve results.}.

%
%


\paragraph{Test Protocol}
At the test time for each image pair, we draw up to 4K minimal (5-point) samples and consider all valid solutions from the solver. Each scoring function with the validated threshold is tasked with selecting the best model incrementally over batches of 100 minimal samples. 
After each 500 samples, we also evaluated the performance of the locally optimized best minimal solution at that point.

\begin{figure}
\centering
\setlength{\tabcolsep}{0pt}
\begin{tabular}{cc}
\multicolumn{2}{c}{\small All Scenes}\\
\includegraphics[width=0.48\linewidth]{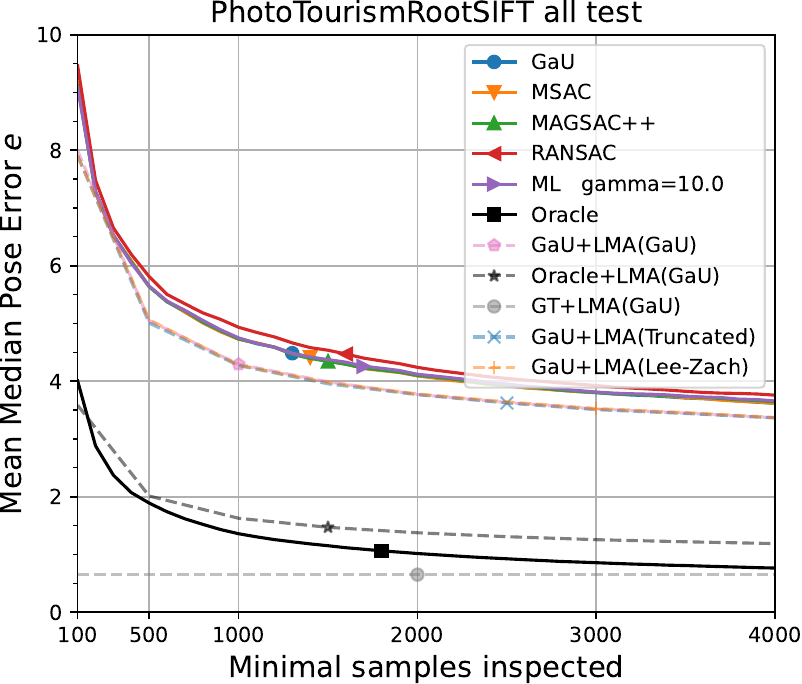}& \ \
\includegraphics[width=0.48\linewidth]{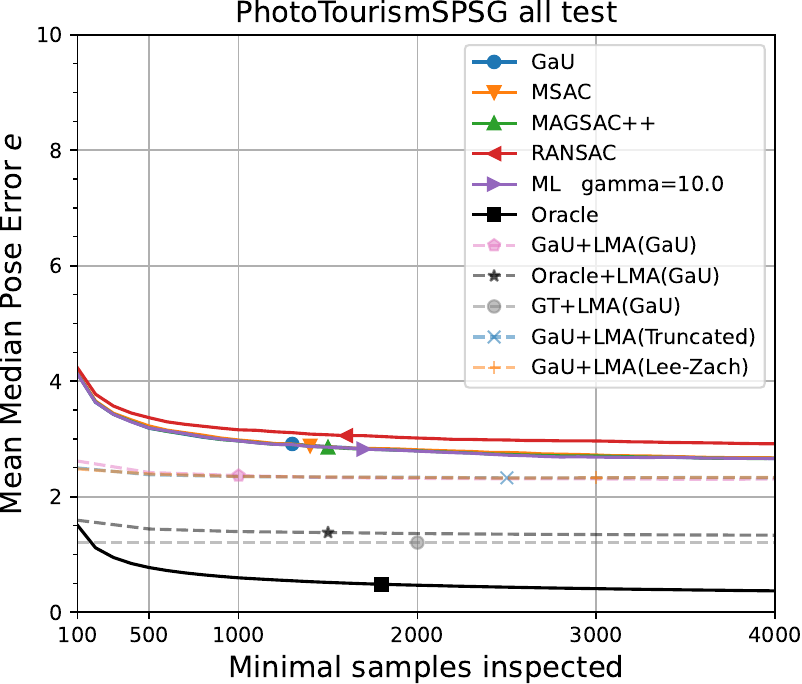}
\end{tabular}
\caption{Pose error of the so-far-the best model selected using different scoring functions while drawing up to 4K minimal samples and scoring all valid solutions. The notation ``X + LMA(Y)'' indicates local optimization using Y, starting from the best minimal solution found so far by X.
 \label{fig:running-all}}
\end{figure}

\paragraph{Results}
\cref{fig:running-all} shows the average performance over all test scenes as a function of the number of samples drawn. It is seen that all scoring methods perform nearly the same, with only RANSAC being marginally worse. Note that this is despite some variation in the shape of the scoring functions in~\cref{fig:xval} (right). This conclusion is confirmed by the detailed metrics in \cref{tab:PhotoTourismRootSIFT,tab:PhotoTourismSPSG}.
Running plots for representative test scenes of varying complexity can be found in \cref{fig:running}.

\begin{figure}[t]
\centering
\setlength{\tabcolsep}{0pt}
\begin{tabular}{cc}
\ \ \ \ \ \ \ {\small Starting with best in 4K samples} & {\small Starting with GT} \ \ \ \ \ \ \ \ \ \ \ \ \ \ \ \ \ \\
\includegraphics[width=0.48\linewidth]{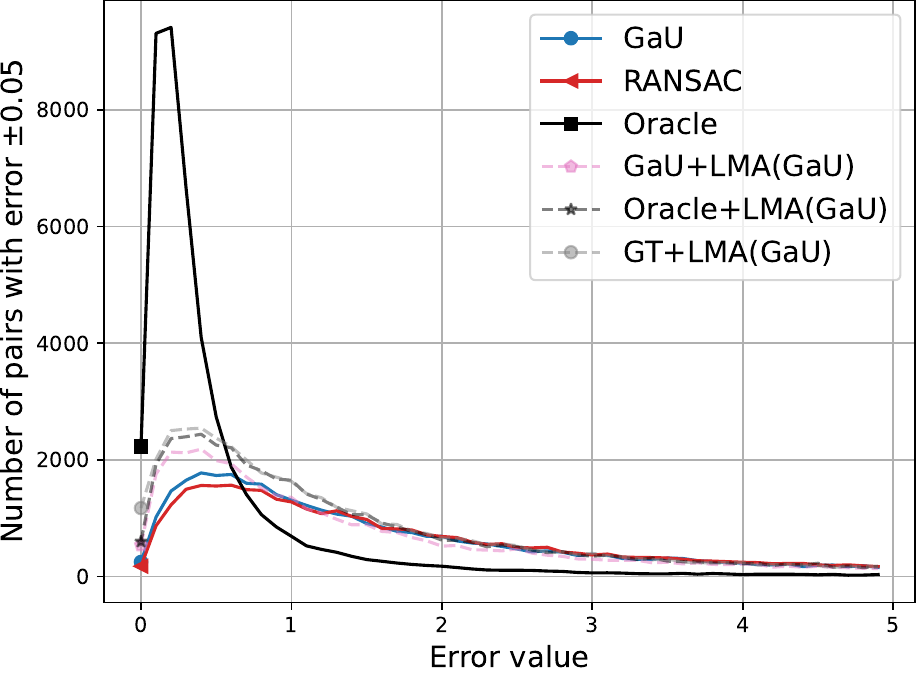}& \ \
\begin{tikzpicture}%
\node[anchor=south west, inner sep=0] (X) at (0,0){%
\includegraphics[width=0.48\linewidth]{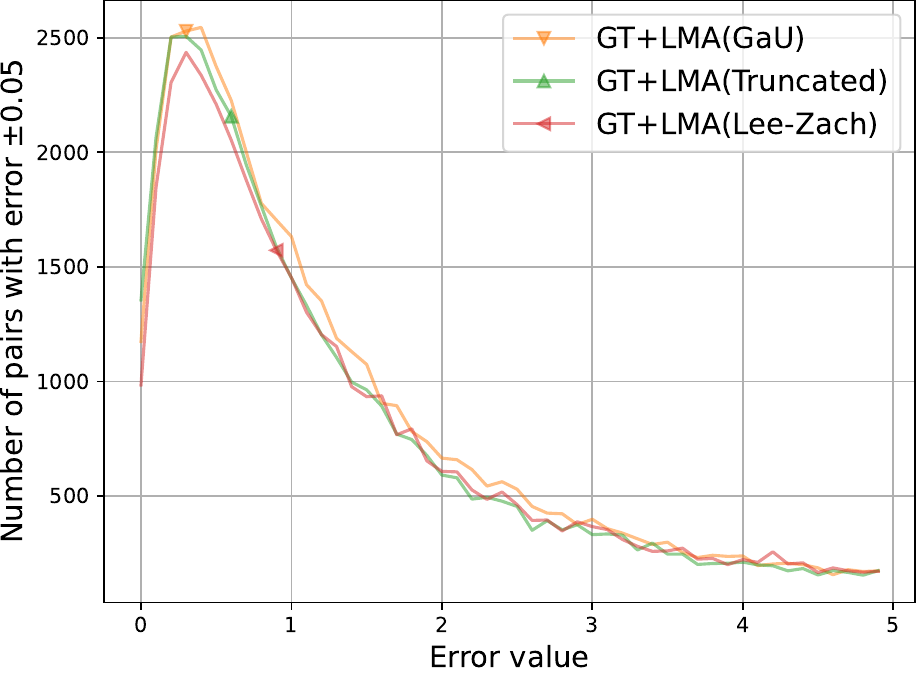}%
};%
\begin{scope}[x={(X.south east)},y={(X.north west)}]%
\node[text width=5cm] (Z) at (0.92,0.5) {%
\resizebox{3cm}{!}{
\setlength{\tabcolsep}{3pt}
\begin{tabular}{rl}
\multicolumn{2}{c}{\small Median}\\
\toprule
GaU: & 1.17 ± 0.016 \\ 
Truncated: & 1.21 ± 0.018\\
Lee-Zach: & 1.32 ± 0.018 \\
\bottomrule
\end{tabular}
}%
};
\end{scope}%
\end{tikzpicture}
\end{tabular}
\caption{Distribution of errors for PhotoTourismSPSG, all test scenes. {\em Left:} Distribution of errors of the best found solution with 4K minimal samples using methods in \cref{fig:running-all}. {\em Right:} distribution of errors using different local optimization techniques when starting from the GT solution.
\label{fig:distribution_err}}
\end{figure}
%
It is clear from \cref{fig:running-all} that it is the number of considered models (and their quality) that plays a crucial role in the final performance rather than the scoring function used.
However, the evaluation with Oracle shows that good models, with relatively low pose error, are actually already available early on, just the scoring functions are not able to select them. We will see how this is explained by geometry in \cref{sec:score-vs-error}.
Local optimization schemes also perform almost the same.  They easily improve results of any valid minimal model when the inlier ratio is high, as seen for the SPSG features easy case in~\cref{fig:running}. However, they cannot substantially improve the poor initial guesses in difficult cases. At the same time, when applied to good initial guesses provided by Oracle, or even the GT model, instead of maintaining the performance, they degrade it. In this case the performance is limited by the information contained in the noisy inlier correspondences.
While we are not competing for the best performance in pose recovery in this paper, we note that the results in~\cref{tab:PhotoTourismRootSIFT,tab:PhotoTourismSPSG} with the final local optimization included are on par with SOTA as discussed in more detail in \cref{A:SOTA}.

\cref{fig:distribution_err} shows the error distribution using different scoring functions, complementing the study in~\cref{fig:running-all} for the median. We see that this distribution is unimodal and rather heavy-tailed, indicating that there is a high number of difficult cases. \cref{fig:distribution_err} (right) evaluates local optimization schemes starting from GT and shows a slightly higher accuracy for the GaU score, which is statistically significant but might be dataset-dependent. 

\subsubsection{Score Consistency and Selectivity}\label{sec:score-vs-error}
In this section, we investigate whether the score (number of inliers or GaU) is or is not indicative of the pose error on average, in order to verify the prior evidence by \citet{barath2022learning}. Specifically, they have observed that almost perfect models have, on average, fewer inliers than the ones that are reasonably accurate but not perfect (\cite[Fig. 2]{barath2022learning}), hinting at a major limitation of the additive scoring and motivating non-linear processing of histograms with MQNet. If that observation were true, it would mean that the score is {\em inconsistent}, \ie better models have systematically lower scores than some worse models, making it impossible to select the best model based on the score.

\begin{figure}[t]
\centering
\setlength{\tabcolsep}{0pt}
\begin{tabular}{cc}
\begin{tabular}{c}\includegraphics[width=0.48\linewidth]{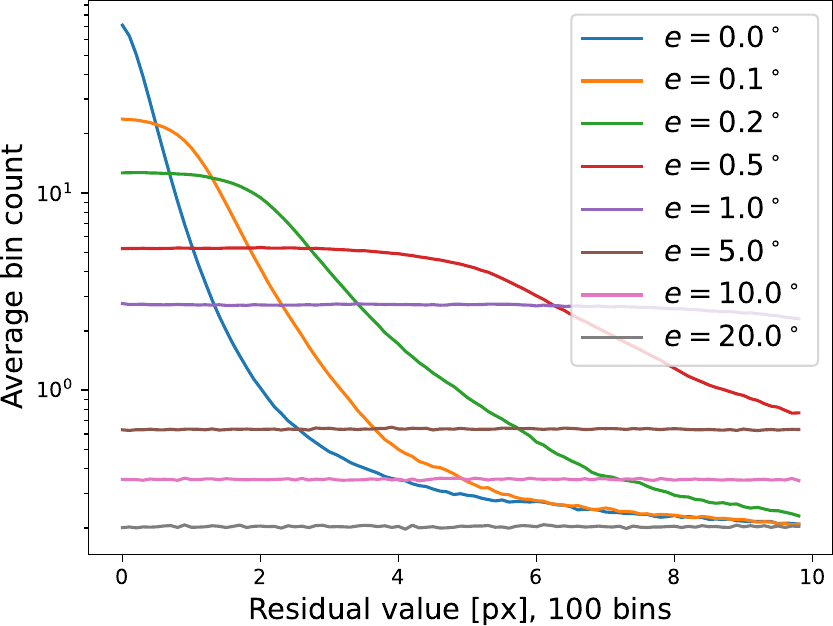}\end{tabular}&\ \ \
\begin{tabular}{c}\includegraphics[width=0.48\linewidth]{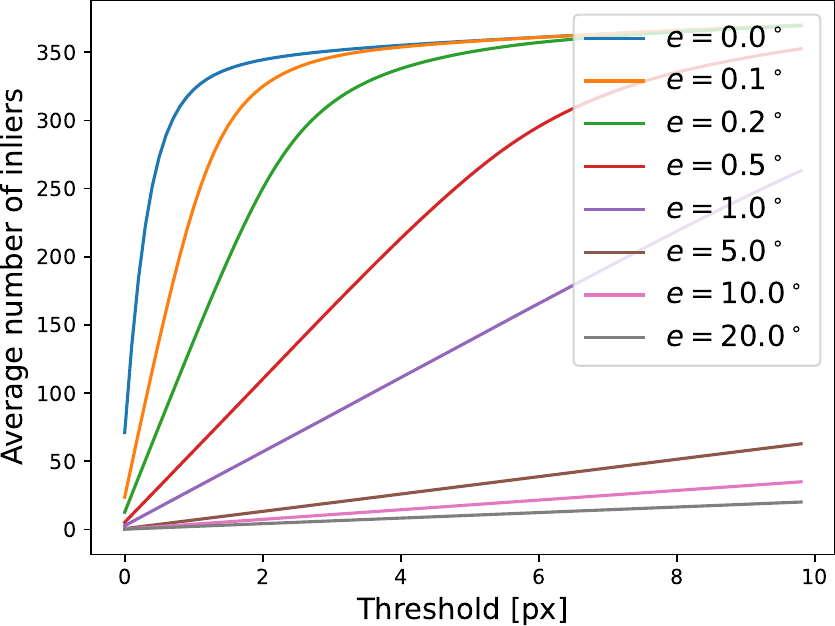}\end{tabular}
\end{tabular}
\caption{Residuals distribution in PhotoTourism RootSIFT with respect to random models at a certain distance from GT measured by pose error. {\em Left}: average count of inliers in each bin (average over all test images).
{\em Right}: number of inliers versus threshold (cumulative sum of bin counts in the left).
\label{fig:residuals_vs_error}
}
\end{figure}

\paragraph{Inliers vs Pose Error}
In \cref{fig:residuals_vs_error} we present a peer verification of the discussed observation. We show average distributions of inliers over all images, where in each image the GT model is randomly rotated (the axis of rotation for $R$ and $t$ is random while the amount of rotation for both is fixed to yield the desired pose error, see \cref{A:exp-inliers}). Contrary to~\cite{barath2022learning} we see that the average number of inliers is higher for more accurate models, for any threshold value. We conclude that the score is {\em consistent} in the sense that the more of correct correspondences are detected (though noisy), the more accurate model we can estimate. We also directly verified that if we add the GT model to a pool of 1K minimal models drawn at random by RANSAC, all scoring methods can quite reliably find it (achieve zero mean median pose error).

The discrepancy to the experiment of \citet{barath2022learning} is likely to be caused by inspecting different statistics. We assume they draw a random image pair\footnote{One issue could be that that the scenes are considered sequentially, according to communication with authors.} and a random minimal model and put it into a bin according to its pose error. This is repeated until each of the pose bins has 10000 models. In this process, the distribution over image pairs (IPs) may not be the same for all bins: \eg we are likely to get in bin “$<0.1$” more models from easier scenes and in bin “$>150$” more models from harder scenes. As a result, we would be comparing statistics in each bin (distributions of residuals) which are not comparable because they are computed {\em for different IP distributions}. Though such a comparison can produce results similar to \citep[Fig. 2]{barath2022learning}, as we have also verified, it has no relevance for RANSAC. RANSAC always compares models for the same IP, and if we want to look at some average statistics, they must be computed over the same distribution of IP.

\begin{figure}[t]
\centering
\setlength{\tabcolsep}{0pt}
\begin{tabular}{cc}
\scriptsize PhotoTourism RootSIFT  & \scriptsize PhotoTourism SPSG \\
\begin{tabular}{c}\includegraphics[width=0.48\linewidth]{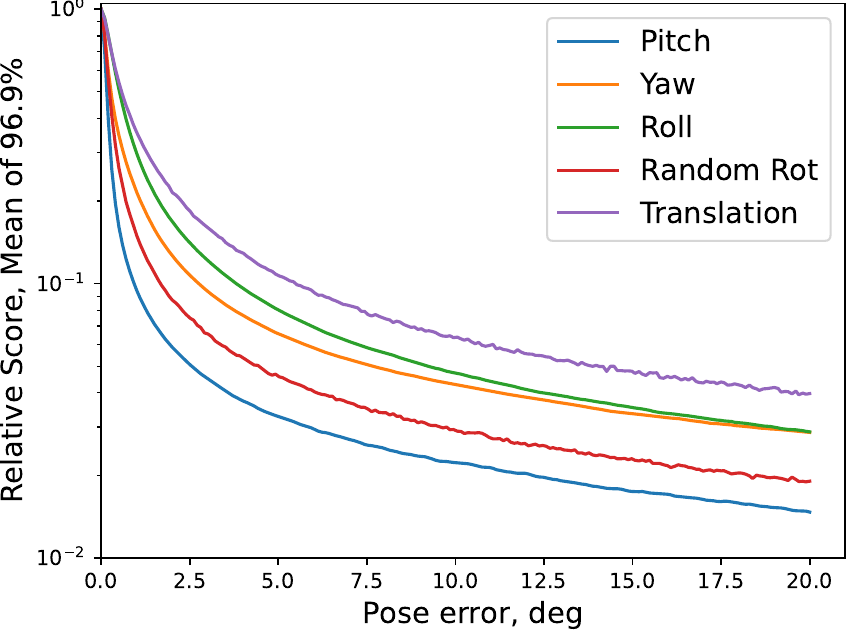}\end{tabular}&\ \ \
\begin{tabular}{c}\includegraphics[width=0.48\linewidth]{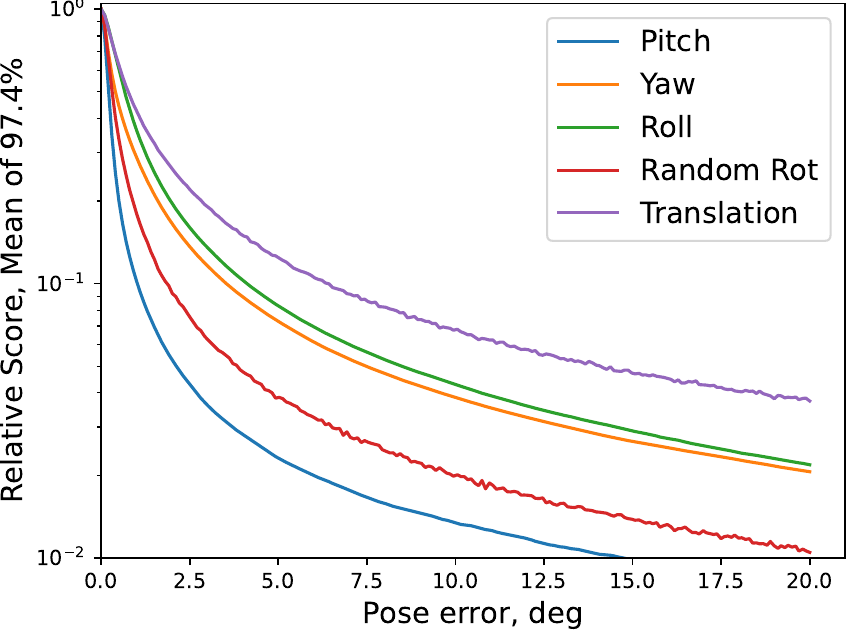}\end{tabular}
\end{tabular}
\caption{Selectivity of the score for deviations of the model from GT along different axis of rotation. The plot is obtained with validated GaU score. Less noisy correspondences in SPSG and tighter threshold lead to a somewhat sharper but qualitatively similar decline.
\label{fig:score_vs_error}
}
\end{figure}

\paragraph{Score Selectivity}
We have verified that the score is consistent. Yet, the Oracle performs significantly better than scoring in~\cref{fig:running-all}. This appears contradictory; however, it can be easily explained. We note that the score has different sensitivities to different degrees of freedom of the relative pose and, as a result, selects models with a small translation or roll error but possibly with a large pitch error. 

Generating random models deviating from GT by a rotation in specific directions (detailed in \cref  {A:exp-inliers}), we obtain \cref{fig:score_vs_error}, which shows the mean relative score versus the rotation angle $\theta$. It is the highest at $\theta=0$ (GT model) and monotonically decreases with all three kinds of rotation: "Pitch", "Yaw", "Roll", albeit at different rates. Apparently, Pitch error leads to a larger change in the residuals per pose degree than the other two. We also show in the same plot rotation of $R$ around a random direction (Random Rot) and a rotation of $t$ around a random direction orthogonal to $t$. Rotation by $\theta$ in all these cases yields pose error of $\theta$. The score is more sensitive to $R$ than $t$, as expected. From this figure, we can see that the Oracle could choose say a model with 2$^\circ$ of Pitch error, but the score would give a preference to a model that may have $10^\circ$ of Roll error or $20^\circ$ of translation error. 

To summarize, we found that the score is on average well-behaved: better models have more inliers and for the same axis of rotation, the score is higher for models closer to the GT. The scatter is still huge and hence we observe significant error levels in practice, but we do not observe any systematic pathologies.
\section{Conclusions}\label{sec:conclusions}

\paragraph{Theoretical Analysis}
This work clarifies and systematizes existing designs of probabilistic and robust scoring functions as well as assumptions underlying these methods. Parametrizing marginal likelihood methods with the threshold enables treating them similarly to profile likelihood and robust methods and allows to see a clear connection between EM and ILRS schemes for local optimizaiton. Using these analysis tools, we have rigorously studied MAGSAC++, arriving at two main conclusions: 1) the derivation is erroneous and these errors / pitfalls, must not be followed by future work, 2) For the relevant degrees of freedom used in practice (MAGSAC++ implementation uses $\nu=4$ throughout for relative pose, homography and 2D line fitting), it is numerically identical to GaU parametric model and cannot theoretically perform better.
More detailed conclusions about MAGSAC++ have been discussed in~\cref{sec:MAGSAC-conclusion}.

\paragraph{Evaluation Methodology}
We have proposed an improved experimental methodology for comparing scoring functions in RANSAC. First, it gives more objective ways of comparing methods. Second, it is efficient for analyzing multiple thresholds simultaneously while computing the inlier likelihood only once. In case when the likelihood is a function of residuals only, a histogram of residuals enables efficient score computation for any set of hyperparameters (not necessarily thresholds) via matrix-matrix product, which is very efficient on GPUs. The fact that the total error metric is aggregated over per-instance errors enables efficient multiple-fold cross-validation.
The methodology is nevertheless limited to evaluating scoring functions and it might be not feasible to adapt in cases such as sequential RANSAC with inner local optimization or multi-model fitting.

\paragraph{Experimental Comparison of Scores}
Our experiments, spanning several estimation problems (H,E,F), datasets and features, and use of the score in minimal model selection and local optimization, confirm not only that MAGSAC++ and GaU are identical in practice, but also that MSAC and the learned scoring function perform similarly in both settings: with a large validation set or with a small validation set in expectation. Only the RANSAC inlier counting is indeed found to be worse. Furthermore, the small validation set experiment reveals that all methods are similarly sensitive to the threshold choice. These observations are in contrast to prior reports in the literature~\cite{barath2021marginalizing,ivashechkin2021vsac,Barath_2023_CVPR} which we attribute to insufficient methodology for evaluation (evaluating whole pipelines, dependence on time budget, dependence on a particular small validation set, dependece on the range and the set of hyperparameters to be validated, unsound definition of sensitivity).
%
\paragraph{Score Consistency}
Our experiments have shown that the scoring functions exhibit consistent behavior: better models (in terms of pose error) are assigned better scores. Thus, there is no discrepancy between the basic probabilistic model and practice. The prior counter-intuitive observation of better models systematically having fewer inliers, motivating using deep model in~\cite{barath2022learning}, is explained by wrong (not relevant) statistics. With the correct experimental methodology, this effect is not observed.


\paragraph{Conclusions for Community}
We thus may conclude that this work clears the ground for future research, both theoretically and experimentally.
The lesson for the research community about MAGSAC++ and MQNet could be as follows. A reliable experimental comparison of {\em methods} (\ie not just of final results on some dataset) could be rather tricky. Authors and reviewers should put more attention to checking that the methods make sense and to the analysis rather than relying on the experimental outcomes as a decisive factor for publication. 
When presenting several improvements to a pipeline, it is necessary to conduct experiments quantifying the main claimed contribution in isolation from other presented improvements as opposed to demonstrating the performance of the complete pipeline. Perhaps the end user cares about the performance of the pipeline, but researchers in the field need to know the merit of the proposed method and which components are important for the performance. 

On the practical side, MAGSAC++ has been demystified. We do not claim that the whole MAGSAC++ pipeline is not better than other methods, but the scoring function is not the reason for this. It can be cleaned up and replaced by the simpler GaU scoring function (which may matter for local optimization), with more interpretable parameters. For changing the setup and or applying to other domains, it could be recommended to adapt GaU score directly.

\paragraph{Limitations and Room for Improvement}
The gap between scoring results and the best model according to Oracle is substantial, but can be to a large extent due to the difference in the metrics: the residuals depend monotonously on the pose error but with different sensitivity to different axes of rotation. This problem can be mitigated by drawing more accurate models, \eg by guided sampling.
Nevertheless, the scoring itself has a room for improvement. 
Possible ways of improvement are naturally related to the assumptions of the basic probabilistic model, which we detailed: spherical inlier noise model same for all correspondences, uniform outliers, \etc. Paradoxically, since we have shown that MAGSAC++ is NOT marginalizing over scales, a promising direction would be to revisit this idea and to perform a proper marginalization over scales or estimate these scales from the underlying image.
The question is which of these limitations is the most significant, so that refining it would result in a sizable improvement of performance. 
Another rather promising direction would be to include dissimilarity of visual descriptors into the model, as compared to relying on pre-filtered featureless correspondences based on SNN ratio.
For handling harder scenes with multiple local minima due to symmetries and repetitions, extending the model to multiple competing hypotheses could be beneficial. 

Improved probabilistic models would have a number of advantages to neural network solutions: they can be lightweight, amenable to LMA optimization for better precision and more adaptive through semi-supervised estimation.
NNs require vast supervised training data, which can be collected only over known reconstructions and may not generalize well to new scenes. Evaluating NN score is substantially more computationally expensive, differentiating NN is costly, and methods such as LMA cannot be applied as they are designed for additive scoring functions.

\section*{Acknowledgment}
The presented research has picked up on an initial practical attempt to learn a better score function with T. Wei, to whom I am thankful for sharing the experience and helping with coding and data.
Even despite the authors of MAGSAC++, D. Barath, J. Matas and J. Noskova, have not found time to participate in this work, I would like to sincerely thank them for discussions, which motivated a more thorough and complete analysis. Their questions have ultimately led to pinpointing the root problems, as well as developing a refined evaluation methodology for small validation sets and for measuring the threshold sensitivity.
Let me also thank the anonymous reviewers for their careful reading of the manuscript and for constructive comments, which have helped to improve the presentation and coverage of the paper.
%
Finally, I gratefully acknowledge the support of the Czech Science Foundation grant GA24-12697S. 

\section*{Data availability}

The PhotoTourism dataset is available from \url{https://www.cs.ubc.ca/research/image-matching-challenge/2020/data/}.

RootSIFT correspondences data (with mutual nearest neighbor check) for PhotoTourism as well as the code to compute them are available from \url{https://github.com/ducha-aiki/ransac-tutorial-2020-data}, titled "Data for epipolar geometry training and validation". 

SPSG correspondences are computed by us using the public implementations of SuperPoint (SP)~\cite{detone2018superpoint} and SuperGlue (SG)~\cite{sarlin2020superglue} as described in the paper.

HEB dataset, including the correspondences, is available from \url{https://github.com/danini/homography-benchmark}.

The code for data processing and reproducing all experiments in the paper will be made publicly available on GitHub. The exact data that our processing has generated can be made available upon a reasonable request.

\bibliography{bib}

\newpage
\appendix
\clearpage

\setcounter{figure}{0}
\setcounter{table}{0}
\counterwithin{figure}{section}
\counterwithin{table}{section}
\def\threshold{\tau}
\section{Proofs}\label{A:proofs}
\TCutoffNew*

\begin{proof}
The residual scoring function $\rho$ of profile likelihood~\eqref{profile-quality} expresses as 
\begin{subequations}
\begin{align}
\rho^\text{\sc prof}(r) & = \log \max \big( \gamma p_{\rm in}(r), (1-\gamma) p_{\rm out}(r) \big)\\
& = \log \max \big( \gamma p_{\rm in}(r), (1-\gamma) \alpha \big)\\
& = \log \max \big( p_{\rm in}(r), \frac{1-\gamma}{\gamma} \alpha \big)+ \log \gamma\\
& = \max (\log p_{\rm in}(r), \log \mu \big) + \log \gamma\\
& = \max (\log p_{\rm in}(r) - \log \mu, 0 \big) + \log \mu\gamma.
\end{align}
\end{subequations}
The normalized form, satisfying $\hat \rho^\text{\sc prof} (T) = 0$ and $\hat \rho^\text{\sc prof}(0) = 1$ is
\begin{subequations}
\begin{align}
\hat \rho^{\text{\sc prof}}(r) = \max \Big(\frac{\log p_{\rm in}(r) - \log \mu}{\log p_{\rm in}(0) - \log \mu}, 0 \Big).
\end{align}
\end{subequations}

Next we show the expression for the marginal likelihood. 
The residual scoring function $\rho$ for marginal-likelihood~\eqref{marg-quality} takes the form
\begin{subequations}
\begin{align}
\rho^\text{\sc marg}(r) & = \log \Big( \gamma p_{\rm in}(r) + (1-\gamma) p_{\rm out}(r) \Big)\\
& = \log \Big( \gamma p_{\rm in}(r) + (1-\gamma)\alpha \Big)\\
& = \log \Big( p_{\rm in}(r) + \frac{1-\gamma}{\gamma}\alpha \Big)  + \log \gamma\\
& = \smax\Big(\log p_{\rm in}(r), \log \mu \Big) + \log \gamma\\
& = \smax\Big(\log p_{\rm in}(r) - \log \mu, 0 \Big) + \log \mu \gamma.
\end{align}
\end{subequations}

The normalized form is: 
\begin{align}
\hat \rho^\text{\sc marg}(r) = \frac{\smax\big(\log p_{\rm in}(r) - \log \mu, 0 \big)}{\smax\big(\log p_{\rm in}(0) - \log \mu, 0 \big)}.
\end{align}
%
\end{proof}

\PRansac*
\begin{proof}
For uniform density of inliers $p_{\rm in}(r) = \beta \bbleft 0 \leq r \leq \frac{1}{\beta}\bbright$. When $\beta>\alpha$, \cref{L:mon} holds with $\tau = \frac{1}{\beta}$. The score ~\eqref{T1Q-prof} expands as
\begin{align}
\rho^{\text{\sc prof}}(r) = \bbleft r < \tau \bbright  (\log \beta - \log\mu).
\end{align}
The normalized score is therefore $\rho^{\text{\sc prof}}(r) = \bbleft r < \tau \bbright$ and $\hat Q^\text{\sc prof} = |\I|$ with $\I$ defined as in\cref{L:mon}.

The marginal residual scoring function for uniform inliers takes the form
\begin{align}
\rho^{\text{\sc marg}}(r) &= \smax(\log p_{\rm in}(r) - \mu, 0)\\
 &= \smax\Big(\log  (\bbleft r \leq \tau \bbright \frac{1}{\tau}) - \mu, 0\Big)\\
& = \begin{cases}
\smax(-\log\tau - \mu, 0) & r \leq \tau,\\
\smax(-\infty, 0) & r > \tau.
\end{cases}
\end{align}
After dividing by $\smax(-\log\tau - \mu, 0) > 0$, we obtain the same result $\rho^{\text{\sc prof}}(r) = \bbleft r < \tau \bbright$.
\end{proof}


\PGaU*

\begin{proof}
For Gaussian density of inliers $p_{\rm in}(r) = \frac{1}{\sqrt{2\gamma} \sigma}e^{-\frac{r^2}{2 \sigma^2}}$ we obtain
\begin{subequations}
\begin{align}
\hat \rho^{\text{\sc prof}}(r) & = \max(\log p_{\rm in}(r) - \log p_{\rm in}(\tau), 0) \\
& = \max\Big(-\frac{r^2}{2 \sigma^2} +\frac{\tau^2}{2 \sigma^2}, 0\Big)	
\propto \max\Big(-\frac{r^2}{\tau ^2} + 1, 0\Big).
\end{align}
\end{subequations}

The marginal residual scoring function takes the form
\begin{align}
\rho^{\text{\sc marg}}(r) &= \smax(\log p_{\rm in}(r) - \log p_{\rm in}(\tau), 0)\\
 &= \smax\Big(\frac{\tau^2 -r^2}{2 \sigma^2 }, 0\Big).
\end{align}
The normalized for $\hat \rho^{\text{\sc marg}}(r) = \rho^{\text{\sc marg}}(r) / \rho^{\text{\sc marg}}(0)$ follows.

The posterior probability of inliers expresses from~\eqref{p_inliers} as
\begin{align}
p(k=1 | r) = \frac{p_{\rm in}(r) }{p_{\rm in}(r) + p_{\rm in}(\tau)} = \frac{e^{-\frac{r^2}{2 \sigma^2}}}{e^{-\frac{r^2}{2 \sigma^2}} + e^{-\frac{\tau^2}{2 \sigma^2}}} = \sigm\Big(\frac{\tau^2-r^2}{2 \sigma^2}\Big).
\end{align}
\end{proof}

\PEMIRLS*
\begin{proof}
\noindent{\bf EM}~  
Given the current model $\theta^t$, The {\bf E}-step of EM, estimates posterior probabilities of points being inliers $p(k_i{=}1 | r_i^t)$. The outliers probabilities are, respectively, $1-p(k_i{=}1 | r_i^t)$. The {\bf M}-step maximizes the lower bound on the marginal likelihood
\begin{align}
\textstyle \sum_{i} p(k_i{=}1 |r_i^t) \log (\gamma p_{\rm in}(r_i)) + p(k_i{=}0 |r_i^t) \log ((1-\gamma) p_{\rm out}(r_i)).
\end{align}
If we assume that $\gamma$ and $p_{\rm out}(r_i) = \alpha$ are fixed and take into account that the weights are fixed by the parameters of the previous iteration, the M step simplifies. After this simplification EM for $\theta$ can be written as:
\begin{subequations}
\begin{align}
& \textstyle r^t_i = r(x_i, \theta^t);\\
& \textstyle q^t_i = p(k_i{=}1 | r_i^t) = \sigm(\log p_{\rm in}(r_i^t) - \log p_{\rm in}(\tau)); \\ 
& \textstyle \theta^{t+1} = \argmax_{\theta} \sum_i q^t_i \log p_{\rm in}(r_i; \theta)).
\end{align}
\end{subequations}

\noindent{\bf IRLS}~ 
In order to apply IRLS to the maximization of $Q(\theta) = \sum_i \rho(r_i)$ we first define $f(R) = \rho(\sqrt{R})$ so that $\rho(r_i) = f(r_i^2)$. We then linearise $f$ as $f(R) \approx f(R^t) + f'(R)(R - R^t)$. Under this approximation 
$\rho(r_i) = f(r_i^2) \approx \rho(r_i^t) + \frac{\rho'(r^t_i)}{r^t_i} (r_i^2 - (r_i^t)^2)$
and IRLS iteration becomes
\begin{align}\label{A:IRLS}
\theta^{t+1} = \arg\min_\theta \sum_i w^t_i r_i^2,
\end{align}
where $r_i = r(x_i, \theta)$ and $w^t_i = -\frac{\rho'(r^t_i)}{r^t_i}$ are the weights. Note that contrary to what the name suggests, this iteration is generally a non-linear weighted least squares problem in $\theta$.
For the basic model we can further express the weights as
\begin{align}
w(r) & = -\frac{\rho'(r)}{r} = -\frac{1}{r}\smax(\log p_{\rm in}(r) - \log p_{\rm in}(\tau), 0)' \\
& = \sigm(\log p_{\rm in}(r) - \log p_{\rm in}(\tau))\frac{1}{r}\frac{d}{d (r)}(-\log p_{\rm in}(r))\\
& = p(k{=}1\mid r) \frac{d}{d (R)}(-\log p_{\rm in}(\sqrt{R}))\Big|_{R = r^2},
\end{align}
which exposes the posterior probability factor $p(k{=}1\mid r)$ and the factor coming from approximating $\log p_{\rm in}$.
\end{proof}

\PGaUEM*
\begin{proof}
We verify that IRLS weights match the posterior inlier probabilities, up to common scale:
\begin{align}
w^{\text{\sc GaU}}(r)  = -\frac{1}{r}\frac{d \rho^{\text{\sc GaU}}(r)}{d r} = -\frac{1}{r}\sigm\Big(\frac{\tau^2 -r^2}{2 \sigma^2 }\Big) \frac{-2r}{2 \sigma^2 } \propto \sigm\Big(\frac{\tau^2 -r^2}{2 \sigma^2 }\Big) = p(k_i{=}1 | r).
\end{align}

Note that the derivation holds under the assumption of constant $\gamma,\alpha$. It is still possible to re-fit $\sigma$ in the M step using the same expression, then $\tau$ becomes a dependent quantity and needs to be updated as
\begin{align}
\tau = \sigma \log\Big(\frac{\gamma}{\sqrt{2 \pi}(1-\gamma)}\frac{1}{\alpha \sigma}\Big).
\end{align}
If the optimization in $\alpha$ or $\gamma$ is desirable as in MLSAC, full marginal likelihood expression should be used\footnote{In MLSAC the inliers threshold $\tau$ ($T$ in \cite{torr2000mlesac}) is set proportional to $\sigma$, \ie it is a fixed quantile of inlier density and $\alpha$ is considered given by the search window size.}.
\end{proof}

\subsection{MAGSAC++}\label{A:MAGSAC++}

\begin{example}[2D Line]\label{A:Line2D}
In the 2D line fitting example of \citet[Sec. 7.4]{barath2021marginalizing}, a true point is sampled on the line uniformly and corrupted by Gaussian noise to produce the observed point with coordinates $(x,y)$, \ie fully consistent with our synthetic model in~\cref{sec:model}.
Without loss of generality, we may assume that the line coincides with the $x$-axis. Then the distance from $(x,y)$ to the line is given by $|y|$. The distribution of $|y|$ is Gaussian (on the positive range, same as the chi distribution with 1 degree of freedom, $\chi_1$). \citet{barath2021marginalizing} assume it is $\chi_2$, a qualitatively different distribution with mode at $|y| = \sigma$, where $\sigma$ is the noise scale, and with zero density at $0$.
\end{example}

\begin{example}[ML under a parameter-dependent transform]\label{example-ML-transform}
Let $X \sim p_X(x; \theta^*)$ be observations, where $\theta^*$ is the true parameter value. In a common setting, we speak of ML estimate of $\theta$ based on the likelihood $p_X(x; \theta)$. Let $Y = f(X; \theta)$ be a parameter-dependent transform of observations (\cf residuals). The density of $Y$ can be found to be (according to the schematic identity $ p_Y(y) dy = p_X(x) dx$):
\begin{align}
p_Y(y; \theta^*, \theta) = p_X(x; \theta^*)  |f'(x; \theta)|^{-1},
\end{align}
where $x = f^{-1}(y; \theta)$ and $f'$ denotes the derivative of $f$ in $x$.
Suppose we write the likelihood of $Y$ as $L_Y(\theta) = \log p_Y(Y ; \theta, \theta)$, \ie the probability to observe $Y$ obtained when using $\theta$ for both: the parameters of the density $p_X$ and of the transform $f$, and consider the estimate maximizing this likelihood:
\begin{align}
\textstyle \hat \theta_Y = \arg\max_{\theta} \prod_i p_Y(y_i; \theta, \theta),
\end{align}
where $r_i=f(x_i, \theta)$ and $x_i$ are i.i.d. from $p_X(x; \theta^*)$. 
Estimate $\hat \theta_Y$ {\em is not a maximum likelihood estimator}. We can see this by inspecting the expected log-likelihood, \ie, the criterion in the limit of an infinite number of i.i.d. observations. It expresses as:
\begin{align}
\mathbb{E}[\log p_Y(y ; \theta, \theta)] & =  \int p_Y(y; \theta^*, \theta) \log p_Y(y ; \theta, \theta) dy\\
& = \int p_X(x; \theta^*) \log \frac{p_X(x; \theta)}{|f'(x; \theta)|} dx 
= \mathbb{E}[\log p_X(x; \theta)] - \mathbb{E}[\log f'(x; \theta)].\notag
\end{align}
If $\log f'(x; \theta)$ is not a constant function of $\theta$, this can easily lead to inconsistency. 
For example, if we sophisticatedly chose $f$ such that $f'(x; \theta) = p_X(x; \theta)$ then $\mathbb{E}[\log p_Y(y ; \theta; \theta)] = 0$ and $\hat \theta_Y$ may be completely arbitrary.

A sound estimation principle in the case of a parameter-dependent transform would be to start from the KL divergence between (a parameter-dependent) observed distribution $p_Y(y; \theta^*, \theta)$ and the model distribution $p_Y(y; \theta, \theta)$. It is easy to verify that this KL divergence (as well as its Monte Carlo estimate from a finite sample) is invariant of any transform, and minimizing it is equivalent to maximizing the likelihood of $X$.
\end{example}

\paragraph{P1: Scale identity for $p_{\rm in}(r; \bar\sigma)$}~\\
Indeed, by making change of variables $\sigma' = \sigma/{\bar \sigma}$ we verify
\begin{align}
p_{\rm in}(r; \bar\sigma) = \frac{1}{\bar \sigma}\int_{0}^{1} \frac{1}{\sigma' \bar \sigma}p_{\bar \chi}\big(\frac{r}{\sigma' \bar \sigma}\big) d \sigma' \bar \sigma = \frac{1}{\bar \sigma}\int_{0}^{1} \frac{1}{\sigma'}p_{\bar \chi}\big(\frac{r}{\sigma' \bar \sigma}\big) d \sigma' = \frac{1}{\bar \sigma}p_{\rm in}(\frac{r}{\bar \sigma}; 1).
\end{align}
The base density $p_{\rm in}(x; 1)$ expresses as
\begin{align}
p_{\rm in}(x; 1) \propto G(x,\kappa,\nu) := \Big(\Gamma\Big(\frac{\nu-1}{2}, \frac{x^2}{2}\Big) - \Gamma\Big(\frac{\nu-1}{2}, \frac{\kappa^2}{2}\Big)\Big)\bbleft x < k\bbright.
\end{align}

\paragraph{P2}~\\
The density $p_{\rm in}$ is incorrectly interpreted as the probability of a point being an inlier 
by defining\footnote{The next equation after (3) in~\cite{barath2021marginalizing}.}
\begin{align}\label{wrong-posterior}
    P(k{=}1 \mid r, \sigma) = g(r \mid \sigma).
\end{align}
This is incorrect because, for example, $p_{\rm in}(r_i)$ might be greater than $1$ as a density. To define the probability of a point being inlier, an outlier model and the prior probability of inliers are necessary, about which the paper makes no assumptions. 

Assume a slightly less restrictive assumption than~\eqref{wrong-posterior}, namely the proportionality:
\begin{align}
    P(k{=}1 \mid r, \sigma) = c g(r \mid \sigma)
\end{align}
for some constant $c$. However, the posterior probability of a point being an inlier and the inlier density are not independent objects. Stating this proportionality requires the existence of some joint distribution of inliers and outliers for which this proportionality can hold. It is straightforward to infer from it that the density of outliers must take the form
\begin{align}
p_{\rm out}(r |\sigma) = \frac{\gamma}{1-\gamma}(\frac{1}{c} - g(r | \sigma) )
\end{align}
for some $\gamma \in (0,1)$. This contradicts the claim of making no assumptions about outliers. Moreover, such outlier density is extremely unreasonable. It is clear that the step~\eqref{wrong-posterior}, being an error in the first place, cannot be reinterpreted as some kind of a reasonable assumption either.

Marginalizing the above defined $P(k{=}1 \mid r, \sigma)$ over $\sigma$,~\cite{barath2021marginalizing} obtain \footnote{Equation (5) in~\cite{barath2021marginalizing}.}
\begin{align}
  P(k{=}1 \mid r) = \frac{1}{\bar \sigma}\int_{0}^{\bar \sigma} g(r \mid \sigma) = p_{\rm in}(r_i).
\end{align}

\paragraph{P3}~\\
From the derivation of IRLS~\eqref{A:IRLS}, the weight and scoring functions are related by the differential equation $\frac{-\rho'(r)}{r} = \tilde w(r)$. 
Solving it for $\rho$ results in the MAGSAC++ score:\footnote{Our definition of $\rho$ differs in sign from~\cite{barath2021marginalizing}.}
\begin{align}
\rho(r; \bar \sigma) = -\int_{0}^{r} x p_{\rm in}(x; \bar \sigma) d x + \text{\it c}.
\end{align}
Using the scale identity for $p_{\rm in}$ and making a substitution $x = x' \bar \sigma$ we obtain the scale identity for $\rho$:
\begin{align}
\rho(r; \bar \sigma) & = -\int_{0}^{r} x \frac{1}{\bar \sigma}p_{\rm in}(\frac{x}{\bar \sigma}; 1) d x
 = -\bar \sigma \int_{0}^{\frac{r}{\bar \sigma}} x' \bar p_{\rm in}(x'; 1) d x' = \bar \sigma \rho(\frac{r}{\bar \sigma}; 1).
\end{align}
The base score function $\rho(x; 1)$ can be expressed through incomplete Gamma functions as well:
\begin{align}\label{MAGSAC-score-gamma}
\rho(x; 1) \propto 2 G(x,k,\nu+2) - r^2 G(x,k,\nu).
\end{align}
\section{Details of Experiments}\label{A:exp_detail}

\subsection{Error Metrics}\label{A:exp-metrics}
\paragraph{Pose Error}
We follow standard performance metrics.
The rotation and translation errors from $(\hat R, \hat t)$ to $(R,t)$ are computed as
\begin{align}
\textstyle e_R(\hat R, R)= \cos^{-1} ( ( \text{tr} ( \hat R R \T) - 1 ) / 2), \ \ \
\textstyle  e_t(\hat t, t) = \cos^{-1} (\frac{\hat t \T t}{\| \hat t \| \| t\|}),
\end{align}
where $(\hat R, \hat t)$ is a decomposition of the found essential / homography matrix, and $(R, t)$ is the true pose.
A decomposition of $E/H$ may have up to 4 solutions. We assume the correct decomposition can be identified. 

The pose error is defined as $e = \max(e_t, e_R)$, measured in degrees. To summarize results over multiple image pairs within a scene we compute median pose error and Average Accuracy at 10 degrees (AA@$10^{\circ}$, detailed below) statistics. As summary statistics over multiple scenes we use the total median for validation and the mean median/AA over scenes for testing, as commonly reported elsewhere.


\paragraph{Average Accuracy}
Mean Average Accuracy (mAA)~\cite{sarlin2020superglue} measures the Area Under the Curve (AUC) of the errors. 
We compute mAA on relative pose errors thresholded at 10$^\circ$ in this paper to evaluate the selected models by the scoring functions of RANSAC. 
For one single threshold, we plot the recall/ratio of the count of the relative pose errors below the threshold, to the total number of estimations.
Then the area is computed as mAA.
We use the implementation of~\cite{barath2022learning}, without binning.

\subsection{Estimation of Inlier Residual Distributions}
To estimate the inlier residual distributions shown in \cref{fig:inliers}, we used the following procedures. To determine the inliers in PhotoTourism dataset we used the reconstructed 3D points and camera poses and the associations between the reconstructed points and the keypoints in the images provided by~\cite{Jin-21}. A keypoint is considered an inlier if the reprojection error of the corresponding 3D point is below 4 pixels. We then computed Sampson error with respect to GT fundamental matrix in pixel coordinates (Sampson error of essential matrix mapped by camera intrinsics). We used all image pairs in the PhotoTourism test set for estimating the inlier distributions.

For homography inliers plot in \cref{fig:inliers}, we used the HPatches dataset~\cite{HPatches}, which provides image pairs with known homographies. We detected SIFT keypoints in both images and matched them using nearest neighbor matching under the GT homography. More specifically, a correspondence is selected as inlier if the distance between the keypoint in the second image and the projection of the keypoint from the first image using the GT homography is below 5 pixels. We then computed Sampson error with respect to GT homography. Hpatches dataset contains illumination and viewpoint change sequences. We used only the viewpoint change sequences for estimating the inlier distributions. We used all pairs of images in all viewpoint sequence.

\subsection{Homography Estimation Experiment}\label{A:exp-HEB}
We have used the Homography Estimation Benchmark (HEB)~\cite{Barath_2023_CVPR} for homography estimation experiments. 
\paragraph{Preprocessing}
First, we filter out the correspondences with SNN ratio above 0.7. They are not used in sampling or score computation.
\paragraph{Sampler} We used the following reasonable sampler to obtain good quality candidate models.
To draw minimal samples we used the 4-point solver implemented in PoseLib. Prior to the solver, we select only 4-touples of correspondences that preserve the orientation of the points in both images, and are not on the same line and no two points are at the same location. The points are drawn at random without replacement with probabilities proportional to $exp(-SNN^2/(2sigma^2))$, where SNN is the second nearest neighbor ratio provided by the dataset and $\sigma=1/8$. The minimal models are immediately rejected if they do not satisfy the chirality constraint for the minimal sample itself.
\paragraph{Metrics}
The dataset contains the GT relative pose but not the GT homography. We decompose the found homography into rotation and translation using OpenCV {\tt decomposeHomographyMat} function and compute rotation to the GT. The translation error is not evaluated in this experiment. In a separate experiment, we have tried to reconstruct the homography from the set of GT inliers available in the dataset and discovered that it has large translation errors (20 degrees), while the rotations were accurate.
\paragraph{Splitting}
The original split into validation and testing contains only 2 validation scenes. In our experiments we have used all 10 scenes and performed cross-validation by selecting 2 random scenes as validation and 8 remaining ones as test and averaging the performance.

\subsection{Relative Pose Experiments}\label{sec:A-PhotoTourism-details}
In all experiments for relative pose estimation, we have used the second nearest ratio filter with threshold 0.9 as \eg in~\cite{barath2022learning}.

\paragraph{Large Validation Set Experiments} The details of this experiment are as follows.
The validation set of PhotoTourism contains 10 scenes (buckingham\_palace, brandenburg\_gate, colosseum\_exterior, grand\_place\_brussels, notre\_dame\_front\_facade, palace\_of\_westminster, pantheon\_exterior, prague\_old\_town\_square, sacre\_coeur, taj\_mahal, trevi\_fountain, westminster\_abbey).
The test set contains 11 scenes (florence\_cathedral\_side, british\_museum, lincoln\_memorial\_statue, london\_bridge, milan\_cathedral, mount\_rushmore, piazza\_san\_marco, reichstag, sagrada\_familia, st\_pauls\_cathedral, united\_states\_capitol).

The training of ML model was conducted on the training scene (st\_peters\_square).
For the validation of threshold hyperparameters in the large validation set setting~\cref{PhotoTourism-val} we used all validation scenes and 1000 random image pairs per scene.
The test in~\cref{sec:PhotoTourism-test} was performed on all image pairs of all test scenes.

\paragraph{Small Random Validation Sets} The details of this experiment are as follows.
From 12 PhotoTourism validation scenes, we created a stratified pool of 10,000 image pairs; and similarly for the test scenes. For each image pair, the pose error was precomputed for 200 thresholds in $[0.1, 10]$. Expected error and variance were estimated as proposed in \cref{sec:small-method} using 1,000 randomly sampled validation sets of size $n$ (taken from validation scenes) and the test used the 10k pairs of the test scenes. This was repeated for different sizes $n=2,4,\dots 1024$.

%
\subsection{Learned Additive Scoring Function (ML)}\label{sec:learned}
%
We can learn the components of the basic probabilistic model from the data and use either marginal or profile log-likelihood as the quality function. 

Let the training data consists of $B$ image pairs with tentative correspondences $x_{i}^b$ and ground true models $\theta^b$, for $b=1\dots B$.
Let $r_{i}^b = r(x_i^b, \theta^b)$ be residuals to the GT models. Whether the correspondences are inliers or outliers is not known. We consider the basic probabilistic mixture model:
\begin{align}
p(r) = \gamma p_{\rm in}(r) + (1-\gamma)p_{\rm out}(r),
\end{align}
where the inliers density will be learned and the outliers density is assumed to be uniform: $p_{\rm out}(r) = \frac{1}{r_{\rm max}}$. The inliers density $p_{\rm in}(r)$ and the proportion of inliers $\gamma$ can be learned by maximum likelihood. However, from the experimental evidence and the understanding of the outliers' nature, we know that the uniform outlier model holds rather poorly. If $p_{\rm in}(r)$ is chosen flexible, the mixture model tends to explain almost all of the data as inliers. To remedy this, we fix the proportion of inliers $\gamma$ to a deliberately lower value. 
\begin{figure}[t]
\centering
\includegraphics[width=0.8\linewidth]{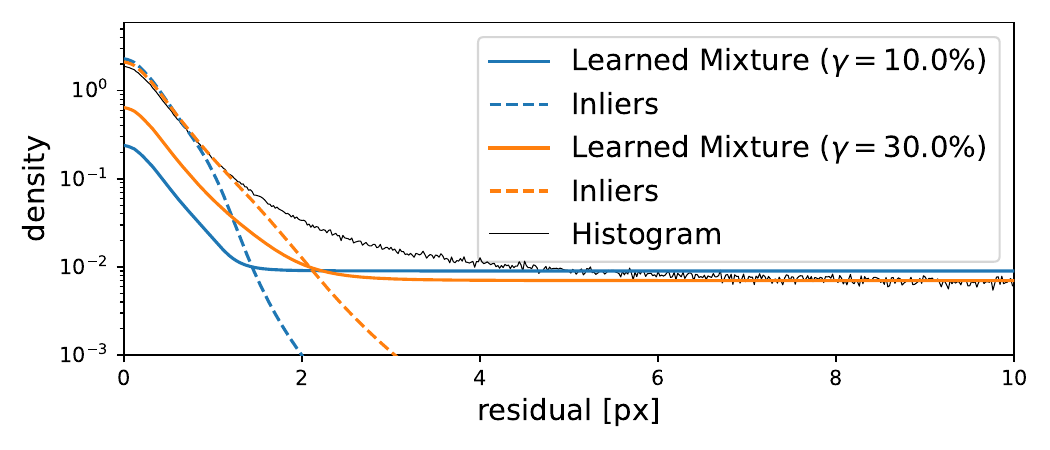}
\caption{Inlier density with the monotonicity constraint learned when fixing $\gamma$, PhotoTourism RootSIFT. \label{fig:learned}}
\end{figure}%

We learn directly the discretized representation of $p_{\rm in}(r)$, \ie, assume it to have the form
\begin{align}
p_{\rm in}(r) = {\rm softmax}\frac{1}{Z}\exp(w^{\rm in}_{k(r)}),
\end{align}
where $w^{\rm in}_k$ are free weights, $k(r)$ is the bin of residual $r$ and $Z$ is the partition function ensuring the proper normalization of this piece-wise constant density.
When using fine discretization and limited training data the estimate may become noisy. We address this by restricting $p_{\rm in}(r)$ to be monotone. This corresponds to the a priori assumption that the density of residuals is higher for smaller residual values. It holds for the considered engineered noise models and corresponds to the common phenomenon that large errors, as a result of a combination of random factors, are rare. We refer to this as {\em monotonic} constraint. The monotonicity constraint on $p_{\rm in}(r)$ can be achieved by representing $w^{\rm in}$ as a cumulative sum of non-negative numbers:
\begin{align}
\textstyle w^{\rm in}_k = \sum_{l>k} \log(1 + e^{\eta_l}).
\end{align}
Here, $\log(1 + e^{\eta_l})$ is non-negative and the reverse cumulative sum over ensures that weights $w^{\rm in}_k$ are monotone decreasing for any choice of $\eta \in \mathbb{R}^K$. We will learn parameters $\eta$. 

The learning only needs residuals to the ground true models $r_i^b$ as the training data. Furthermore, since our likelihood probabilistic model is discretized, to evaluate the loss (log-likelihood of $r_i^b$) during the learning it is sufficient to know only the histogram $H$ of all GT residuals of the whole training set. Once this statistics is computed, the learning is very fast. In fact, one correspondences are known, the estimation of $H$ for 5k image pairs and learning takes just a few seconds. The examples of learned inlier densities are shown in~\cref{fig:learned}.

It remains to discuss how the final score weights $\vw$ are computed once we learned inliers distribution representation $w^{\rm in}$. We use the marginal likelihood quality model and define
\begin{align}
\textstyle w_k = \log (\gamma \frac{1}{Z}e^{w^{\rm in}_k} + (1-\gamma )\frac{1}{r_{\rm max}}),
\end{align}
which are then normalized to have minimum $0$ and maximum $1$. The inlier prior $\gamma$, considered here a hyperparameter, can be represented by the equivalent threshold parameter $\tau$ and tuned similarly to other methods. 

In the experiments in the main paper, we used $r_{\rm max}=100$ and the same range of the effective threshold hyperparameter $\tau$ as for engineered methods. 
\subsection{Score Consistency and Selectivity}\label{A:exp-inliers}
\paragraph{Sampling of Random Rotations}
In~\cref{sec:score-vs-error} we sample rotations around a random axis. The axis direction is sampled by drawing $d \in \Real^3$ with independent standard normal components and normalizing, $d/\|d\|$. This ensures a uniform distribution on the sphere. The amount of rotation in our tests is always prescribed. Let the corresponding rotation matrix be denoted $Q$. Given a GT model $(R,t)$, applying Q to rotation means forming $(Q R, t)$, to translation $(R, Q t)$, and then composing the corresponding essential matrix. Pitch rotation corresponds to a rotation matrix Q rotating by $\theta$ around X axis, Yaw around Y and Roll around Z.

\paragraph{Score Selectivity}
The experiment in~\cref{fig:score_vs_error} is conducted as follows.
First we filter out image pairs with a score of the GT model less than 5 (the score is normalized and can be informally interpreted as a soft inlier count). These constitute $3.1\%$ in RootSIFT and $2.6\%$ in SPSG and correspond to image pairs where we have not enough inlier correspondences for any reasonable estimation.
Starting from a GT model $(R,t)$ we create a deviated model $(R',t')$ where $R'$ differs from $R$ in rotation along a single axis by an angle $\theta$. The pose error of $(R',t')$ is thus $\theta$. We then compute the score of this model relative to the score of the GT model.

\section{Additional Experimental Results}\label{A:experiments}

\subsection{Additional Plots for PhotoTourism Validation and Testing}
\begin{figure}
\centering
\setlength{\tabcolsep}{0pt}
\begin{tabular}{cc}
\multicolumn{2}{c}{PhotoTourism SPSG} \\
\scriptsize Validation  & \scriptsize Selected Kernels \\
\begin{tabular}{c}\includegraphics[width=0.48\linewidth]{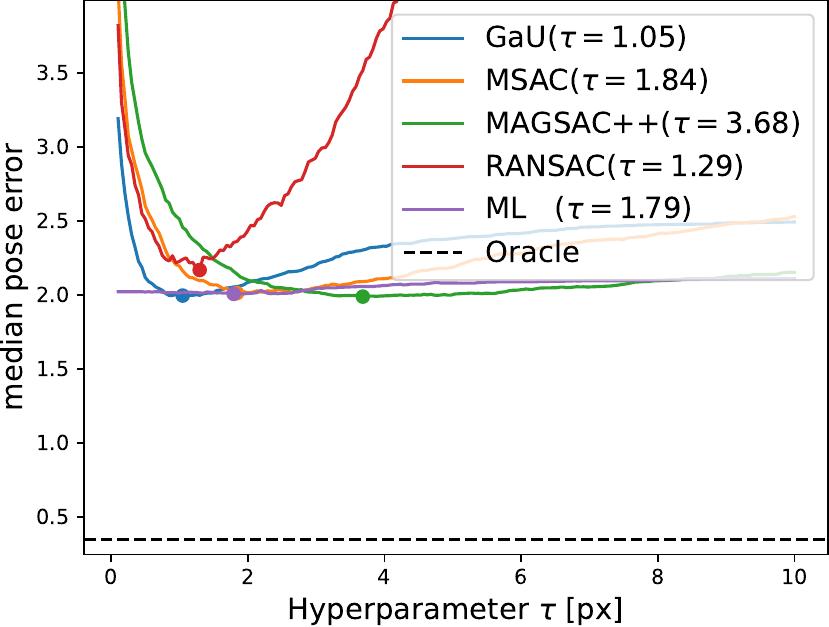}\end{tabular}&
\begin{tabular}{c}\includegraphics[width=0.48\linewidth]{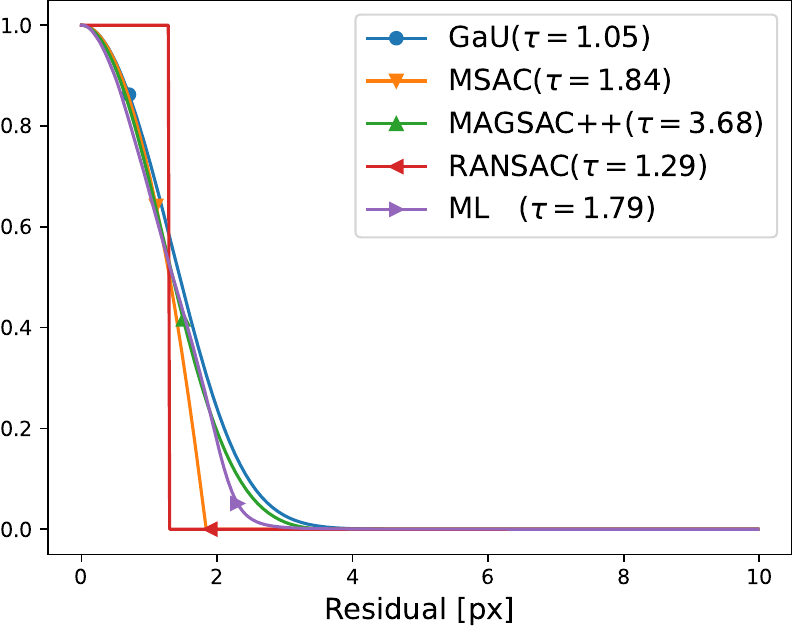}
\end{tabular}
\end{tabular}
\caption{Validation of the threshold hyperparameter on PhotoTourism SPSG. {\em Left}: median error versus threshold. {\em Right:} Scoring functions corresponding to selected thresholds. 
\label{fig:validation_SPSG}
}
\centering
\begin{tabular}{cc}
\multicolumn{2}{c}{PhotoTourism SPSG} \\
\includegraphics[width=0.49\linewidth]{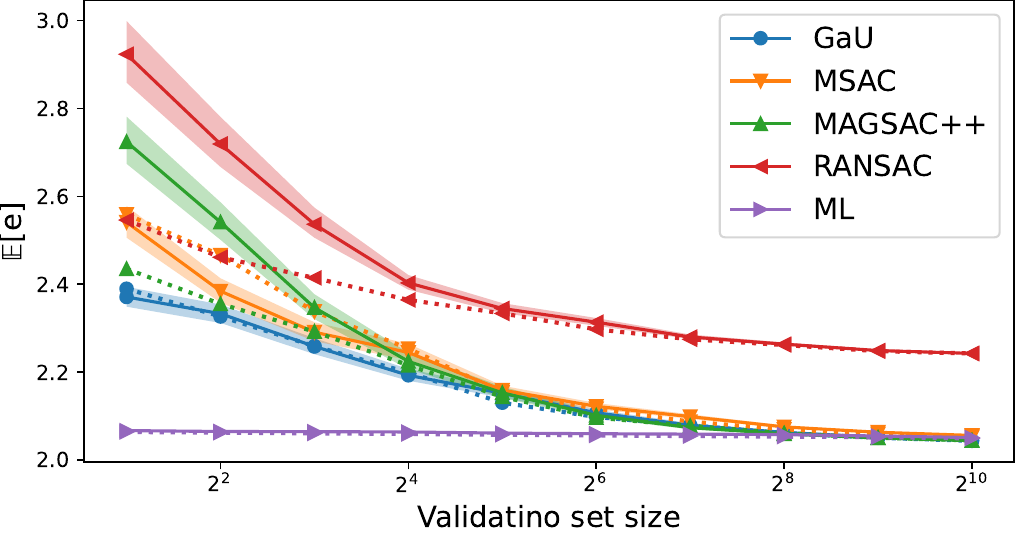}&\ \ 
\includegraphics[width=0.49\linewidth]{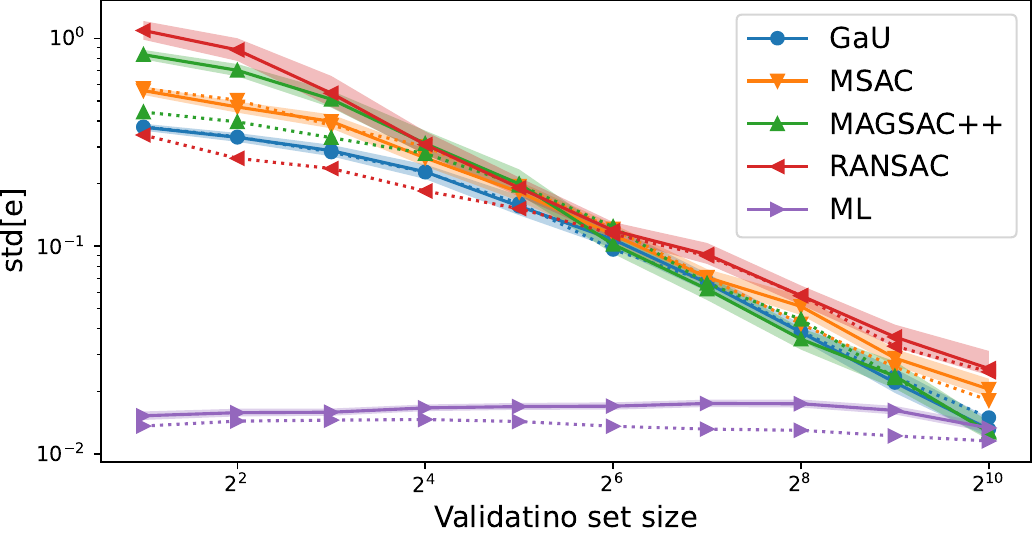}
\end{tabular}
\caption{Small random validation set experiment. Expected test pose error ({\em left}) and its std ({\em right}) in dependence on the validation set size $n$ for PhotoTourism SPSG.
\label{fig:expectd_SPSG}}
\end{figure}

\cref{fig:validation_SPSG,fig:expectd_SPSG} shows validation plots, expected test error and threshold sensitivity for PhotoTourism SPSG features, complementing \cref{fig:xval,fig:expectd_RootSIFT} for RootSIFT features in the main paper. Similar trends are observed: GaU and MAGSAC++ scoring functions are very similar, and all methods exhibit similar sensitivity to the threshold choice, with ML excepted because it has used an extra training scene prior to validation. 

\begin{table}[t]
\centering
\caption{Evaluation on PhotoTourism RootSIFT, 4K minimal samples.
The ``Median'' metric reports mean median of rotation ($e_R$), translation ($e_t$) or combined $e = \max(e_R, e_t)$, where the mean is over 11 test scenes.
The mAA metric reports mean Average Accuracy@10$^{\circ}$. 
The confidence intervals on the metrics are \wrt the test set and are computed with bootstrap 'BCa' with $95\%$ confidence.
\label{tab:PhotoTourismRootSIFT}
}
\begin{tabular}{c}
\resizebox{\linewidth}{!}{
\begin{tabular}{llll|lll}
\toprule
\multicolumn{7}{c}{Minimal Samples Scoring}\\
\midrule
         & \multicolumn{3}{c}{mAA} & \multicolumn{3}{c}{Median} \\
Scoring: & \multicolumn{1}{c}{$e$} & \multicolumn{1}{c}{$e_R$} & \multicolumn{1}{c}{$e_t$} & \multicolumn{1}{c}{$e$} & \multicolumn{1}{c}{$e_R$} & \multicolumn{1}{c}{$e_t$} \\
\midrule
RANSAC & 0.581$\pm$0.01 & 0.739$\pm$0.01 & 0.599$\pm$0.01 & 3.76$\pm$0.25 & 1.54$\pm$0.12 & 3.29$\pm$0.14 \\
MSAC & 0.591$\pm$0.01 & 0.744$\pm$0.01 & 0.610$\pm$0.01 & 3.61$\pm$0.24 & 1.51$\pm$0.11 & 3.18$\pm$0.21 \\
MAGSAC++ & 0.592$\pm$0.01 & 0.744$\pm$0.01 & 0.610$\pm$0.01 & 3.64$\pm$0.23 & 1.53$\pm$0.11 & 3.19$\pm$0.21 \\
GaU & 0.592$\pm$0.01 & 0.744$\pm$0.01 & 0.611$\pm$0.01 & 3.62$\pm$0.23 & 1.52$\pm$0.11 & 3.18$\pm$0.21 \\
Learned & 0.591$\pm$0.01 & 0.744$\pm$0.01 & 0.610$\pm$0.01 & 3.66$\pm$0.23 & 1.53$\pm$0.11 & 3.22$\pm$0.21 \\
Oracle & 0.827$\pm$0.01 & 0.860$\pm$0.01 & 0.849$\pm$0.01 & 0.76$\pm$0.05 & 0.52$\pm$0.03 & 0.64$\pm$0.15 \\
\midrule
\multicolumn{7}{c}{Local Optimization of the Best GaU Minimal Model}\\
LO: & \multicolumn{1}{c}{$e$} & \multicolumn{1}{c}{$e_R$} & \multicolumn{1}{c}{$e_t$} & \multicolumn{1}{c}{$e$} & \multicolumn{1}{c}{$e_R$} & \multicolumn{1}{c}{$e_t$} \\
\midrule
IRLS-LMA GaU & 0.609$\pm$0.01 & 0.755$\pm$0.01 & 0.625$\pm$0.01 & 3.37$\pm$0.23 & 1.38$\pm$0.12 & 2.98$\pm$0.22 \\
PoseLib (Truncated) & 0.609$\pm$0.01 & 0.755$\pm$0.01 & 0.626$\pm$0.01 & 3.36$\pm$0.22 & 1.39$\pm$0.11 & 2.96$\pm$0.21 \\
PoseLib (Le-Zach) & 0.608$\pm$0.01 & 0.755$\pm$0.01 & 0.625$\pm$0.01 & 3.37$\pm$0.23 & 1.39$\pm$0.12 & 2.98$\pm$0.21 \\
\bottomrule
\end{tabular}
}
\end{tabular}
\vskip 1\baselineskip
\caption{Evaluation on PhotoTourism SPSG, 4K minimal samples. \label{tab:PhotoTourismSPSG}
}
\centering
\begin{tabular}{c}
\resizebox{\linewidth}{!}{
\begin{tabular}{llll|lll}
\toprule
\multicolumn{7}{c}{Minimal Samples Scoring}\\
\midrule
         & \multicolumn{3}{c}{mAA} & \multicolumn{3}{c}{Median} \\
Scoring: & \multicolumn{1}{c}{$e$} & \multicolumn{1}{c}{$e_R$} & \multicolumn{1}{c}{$e_t$} & \multicolumn{1}{c}{$e$} & \multicolumn{1}{c}{$e_R$} & \multicolumn{1}{c}{$e_t$} \\
\midrule
RANSAC & 0.633$\pm$0.01 & 0.797$\pm$0.01 & 0.650$\pm$0.01 & 2.92$\pm$0.17 & 1.12$\pm$0.06 & 2.67$\pm$0.16 \\
MSAC & 0.651$\pm$0.01 & 0.807$\pm$0.01 & 0.667$\pm$0.01 & 2.67$\pm$0.15 & 1.04$\pm$0.06 & 2.43$\pm$0.14 \\
MAGSAC++ & 0.653$\pm$0.01 & 0.808$\pm$0.01 & 0.669$\pm$0.01 & 2.66$\pm$0.15 & 1.04$\pm$0.06 & 2.42$\pm$0.15 \\
GaU & 0.653$\pm$0.01 & 0.808$\pm$0.01 & 0.669$\pm$0.01 & 2.66$\pm$0.14 & 1.04$\pm$0.06 & 2.43$\pm$0.14 \\
Learned & 0.653$\pm$0.01 & 0.808$\pm$0.01 & 0.669$\pm$0.01 & 2.66$\pm$0.15 & 1.03$\pm$0.06 & 2.42$\pm$0.14 \\
Oracle & 0.919$\pm$0.00 & 0.932$\pm$0.00 & 0.931$\pm$0.00 & 0.37$\pm$0.01 & 0.27$\pm$0.01 & 0.31$\pm$0.01 \\
\midrule
\multicolumn{7}{c}{Local Optimization of the Best GaU Minimal Model}\\
LO: & \multicolumn{1}{c}{$e$} & \multicolumn{1}{c}{$e_R$} & \multicolumn{1}{c}{$e_t$} & \multicolumn{1}{c}{$e$} & \multicolumn{1}{c}{$e_R$} & \multicolumn{1}{c}{$e_t$} \\
\midrule
IRLS-LMA GaU & 0.678$\pm$0.01 & 0.820$\pm$0.01 & 0.693$\pm$0.01 & 2.30$\pm$0.15 & 0.90$\pm$0.05 & 2.08$\pm$nan \\
PoseLib (Truncated) & 0.677$\pm$0.01 & 0.819$\pm$0.01 & 0.692$\pm$0.01 & 2.33$\pm$0.14 & 0.92$\pm$0.06 & 2.12$\pm$0.14 \\
PoseLib (Le-Zach) & 0.677$\pm$0.01 & 0.819$\pm$0.01 & 0.692$\pm$0.01 & 2.33$\pm$0.14 & 0.91$\pm$0.05 & 2.12$\pm$0.14 \\
\bottomrule
\end{tabular}
}
\end{tabular}
\end{table}

\subsection{Quantitative Comparison to SOTA}\label{A:SOTA}
\begin{figure}[p]
\centering
\setlength{\tabcolsep}{0pt}
\setlength{\figwidth}{0.47\linewidth}
\begin{tabular}{cc}
\multicolumn{2}{c}{Easier Scene: St'Pauls Cathedral}\\[5pt]
\includegraphics[width=\figwidth]{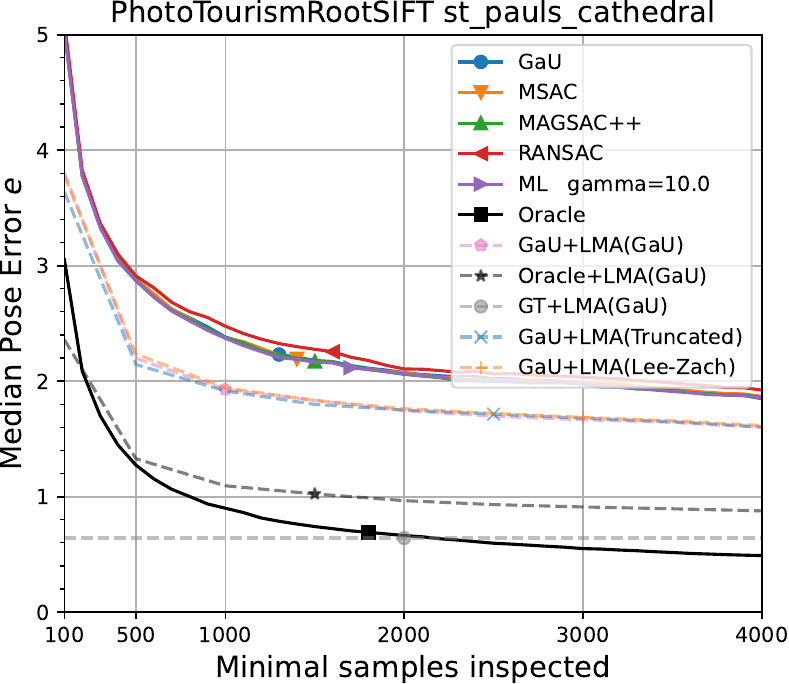}& \ \
\includegraphics[width=\figwidth]{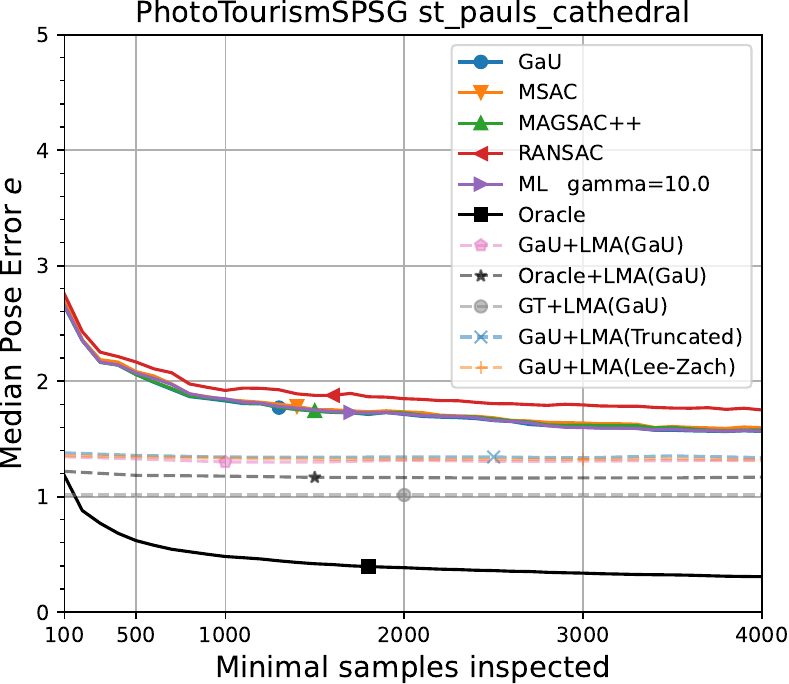}\\[6pt]
\multicolumn{2}{c}{Difficult Scene: London Bridge (two symmetrc towers)}\\[5pt]
\includegraphics[width=\figwidth]{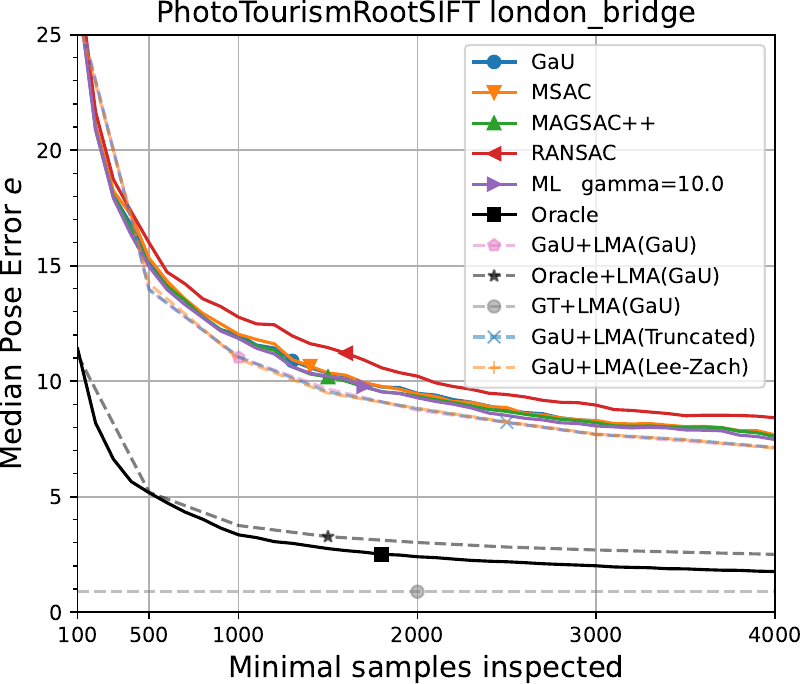}& \ \
\includegraphics[width=\figwidth]{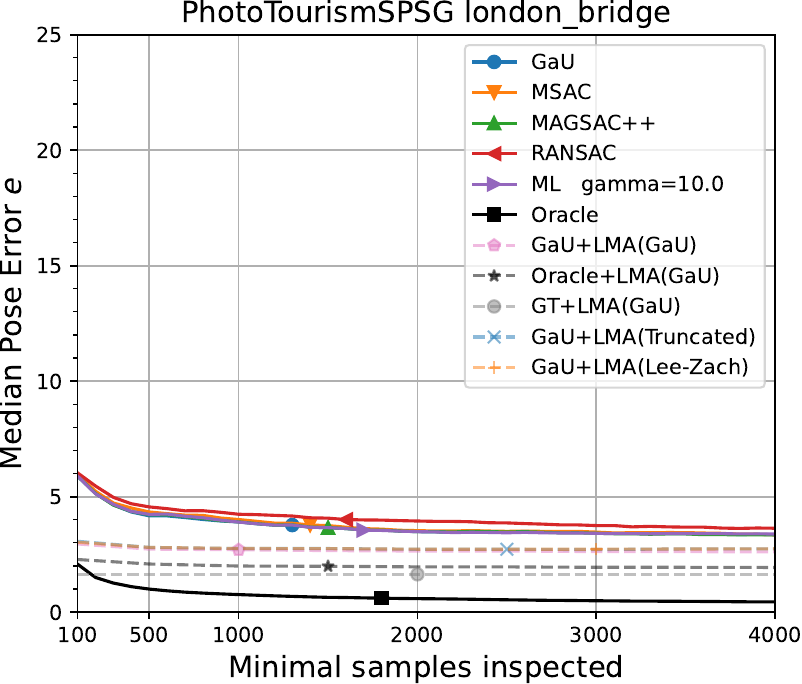}\\[5pt]
\multicolumn{2}{c}{More Difficult Scene: United States Capitol (Repeated Elements + Symmetries)}\\[6pt]
\includegraphics[width=\figwidth]{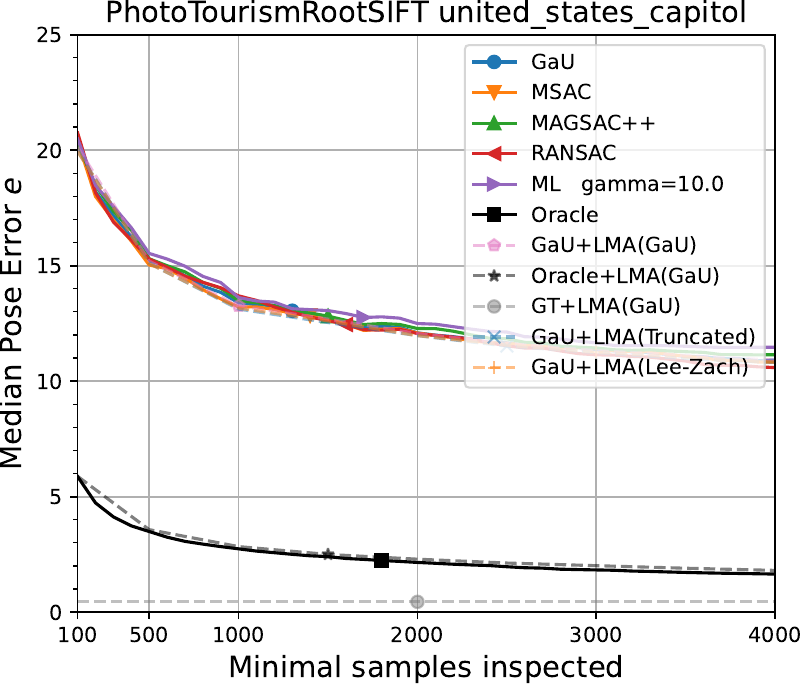}& \ \
\includegraphics[width=\figwidth]{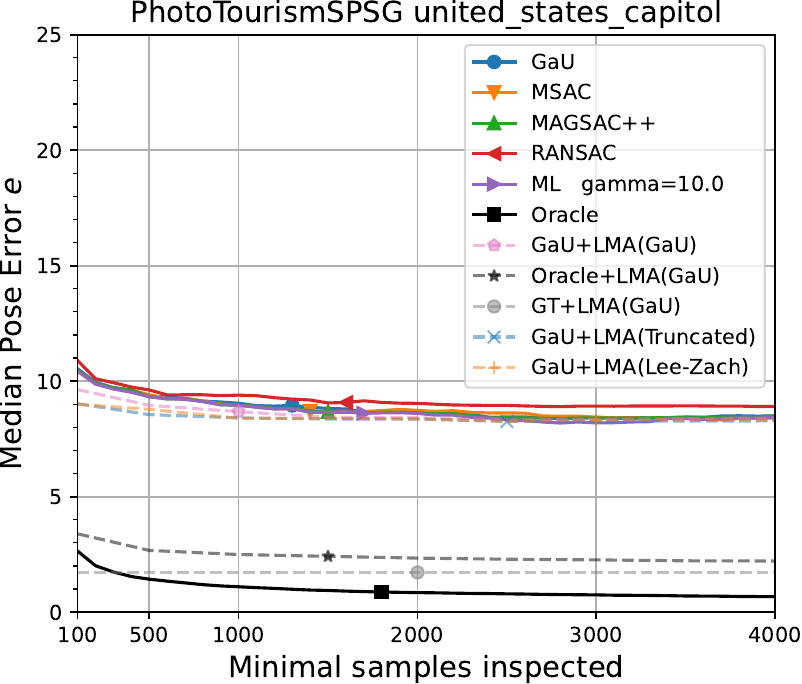}
\end{tabular}
\caption{Pose error of the so-far-the best model selected using different scoring functions while drawing up to 4K minimal samples and scoring all valid solutions. The best model found by GaU is also optimized with local optimization schemes.
The average over all test scenes is given in~\cref{fig:running-all}.
 \label{fig:running}}
\end{figure}

\cref{tab:PhotoTourismRootSIFT} and \cref{tab:PhotoTourismSPSG} report detailed quantitative results on PhotoTourism RootSIFT and SPSG features, complementing the running plots in \cref{fig:running} and \cref{fig:running-all}. We can compare these results and our findings to SOTA results obtained with the same correspondences.

We have not observed a substantial improvement of MAGSAC++ over MSAC. This is consistent with~\cite[Fig.4]{barath2022learning} where MSAC and MAGSAC++ are compared under equal post-processing.
For PhotoTourism RootSIFT, our mAA using 4K samples with LMA for $R$ and $t$  of (0.755, 0.625) in \cref{tab:PhotoTourismRootSIFT} are better than (0.71, 0.47) reported for MAGSAC++ in~\cite[Tab. 2]{barath2022learning} and not worse than (0.76, 0.61) for comparable to MQNet(E)~\cite[Tab. 2]{barath2022learning}\footnote{Table 2 seems to select the best result from the ablation in~\cite[Tab. 3]{barath2022learning}, which suggests that the ablation (selecting amongst different networks for different bin sizes) is performed on the test set. Thus, the best result from the networks that are differently trained is selected based on the test set.}.

MQNet~\cite{barath2022learning} learns a non-linear function of the histogram (hence a non-additive function of residuals), which can potentially break the limitations of additive scoring. But it also includes a complex post-processing (polishing) of the best minimal model, sweeping 10 inlier thresholds, and calling OpenCV {\tt findEssentialMat} on each candidate inlier set, which performs a RANSAC search with an upper bound of 10K iterations. The so-found complex hypotheses are then re-scored by MQNet, followed by LMA.  
Their mean median errors w/o polishing~\cite[Tab. 6]{barath2022learning} of $e_R = 2.22$ and $e_t = 4.47$ are substantially worse than ours even without LMA ($1.52$, $3.18$). Yet median results with polishing $e_R = 0.91$ and $e_t = 2.34$ are substantially better. Since the selection is always done with MQNet, this suggests that the MQNet score is better than MSAC/MAGSAC++ while the inferior 'w/o' result was possibly due to an insufficient number of samples. Unfortunately, there is no direct evidence in~\cite{barath2022learning} comparing MQNet and MSAC/MAGSAC++ scoring for the same post-processing\footnote{In particular, in \cite[Fig.4]{barath2022learning} MSAC and MAGSAC++ use LO post-processing while MQNet uses the proposed complex polishing post-processing.}. To be fair, it should be also noted that if the MQNet scoring is better, then the final LMA, based on MSAC scoring, might actually be degrading the results.

\subsection{Other Datasets, E estimation}\label{A:ETH3D_LAMAR}
\begin{figure}
\centering
\setlength{\tabcolsep}{0pt}
\begin{tabular}{cc}
\includegraphics[width=0.48\linewidth]{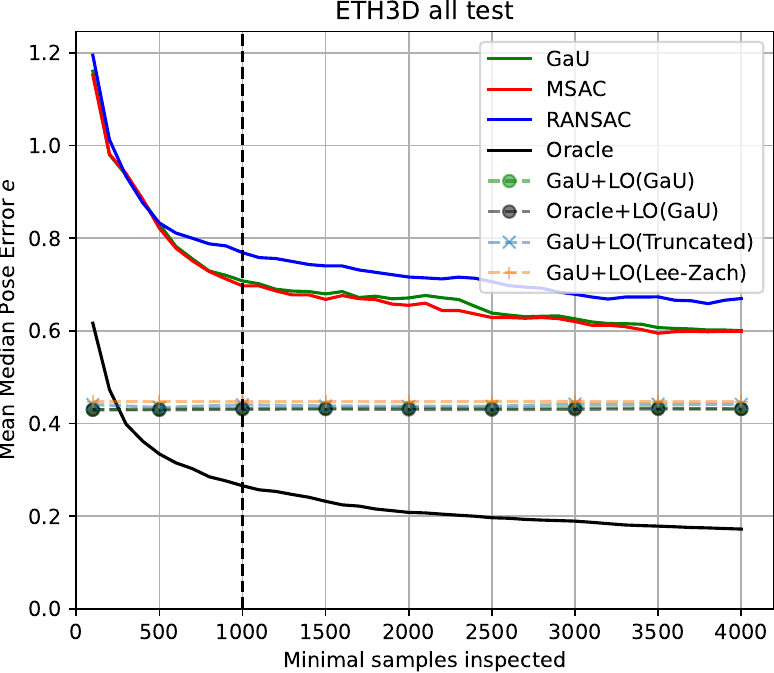}& \ \ 
\includegraphics[width=0.48\linewidth]{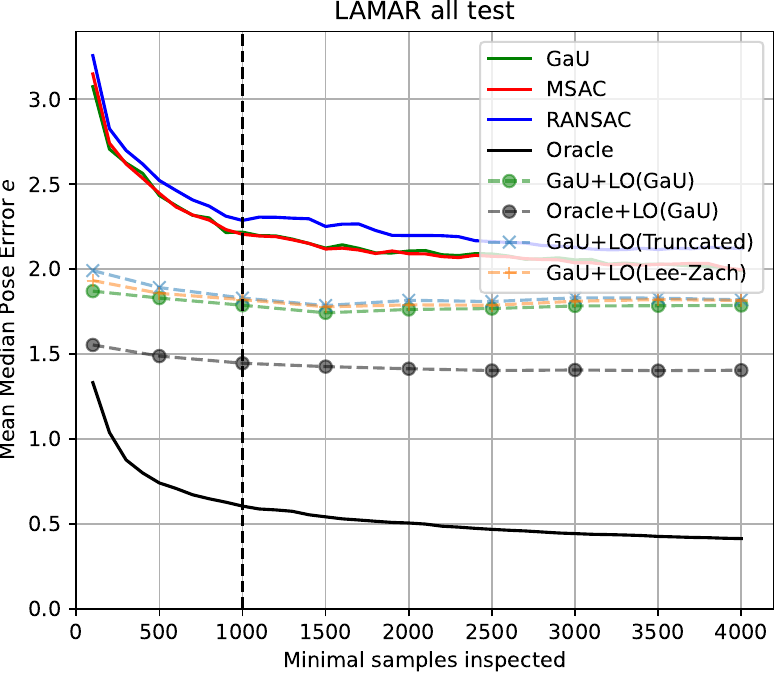}
\end{tabular}
\caption{ETH3D and LAMAR datasets. Pose error of the so-far-the best model selected using different scoring functions while drawing up to 4K minimal samples and scoring all valid solutions.
 \label{fig:LAMAR}}
\end{figure}

\paragraph{ETH3D, LAMAR}
ETH3D appears easy: local optimization quickly snaps the results to the best achievable with any number of minimal samples. The Oracle models are degraded to almost the same level, which indicates we are close to the global optimum of the score function. A further improvement of accuracy is possible only by improving the scoring function and not \eg sampling strategy.
\cref{fig:LAMAR} shows running plots for LAMAR and ETH3D datasets. 

\begin{figure}[p]
\centering
\setlength{\tabcolsep}{0pt}
\begin{tabular}{cc}
\scriptsize Validation  & \scriptsize Selected Kernels \\
\begin{tabular}{c}\includegraphics[width=0.48\linewidth]{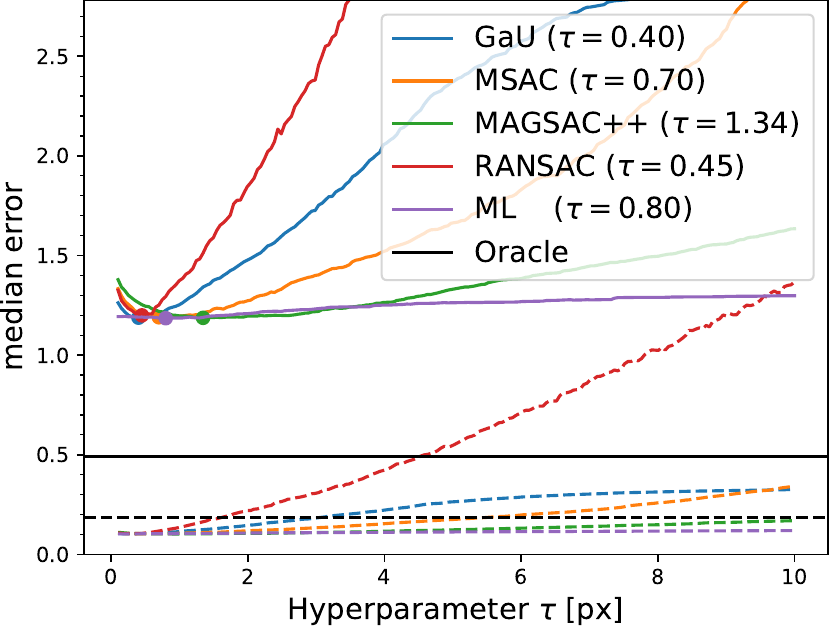}\end{tabular}&\ \
\begin{tabular}{c}\includegraphics[width=0.46\linewidth]{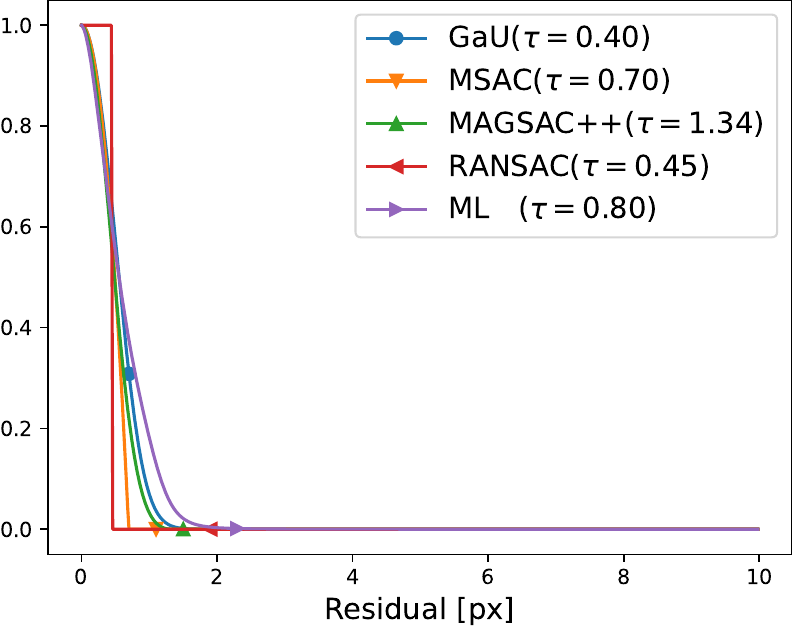}\end{tabular}\\[-5pt]
\end{tabular}
\caption{KITTI F: Validation of the threshold hyperparameter using 7-point F-solver (up to 3 solutions), 4K samples, 5K image pairs.
{\em Left}: median error $e$ (solid) and median rotation error $e_R$  (dashed) versus threshold. The optimal threshold is chosen based on $e$. Clearly $e_t$ dominates the choice but luckily $e_R$ is also close to optimum for the selected thresholds.
{\em Right:} Scoring functions corresponding to selected thresholds.
\label{fig:KITTI-val}
}
\vspace{10pt}
\begin{tabular}{cc}
\includegraphics[width=0.48\linewidth]{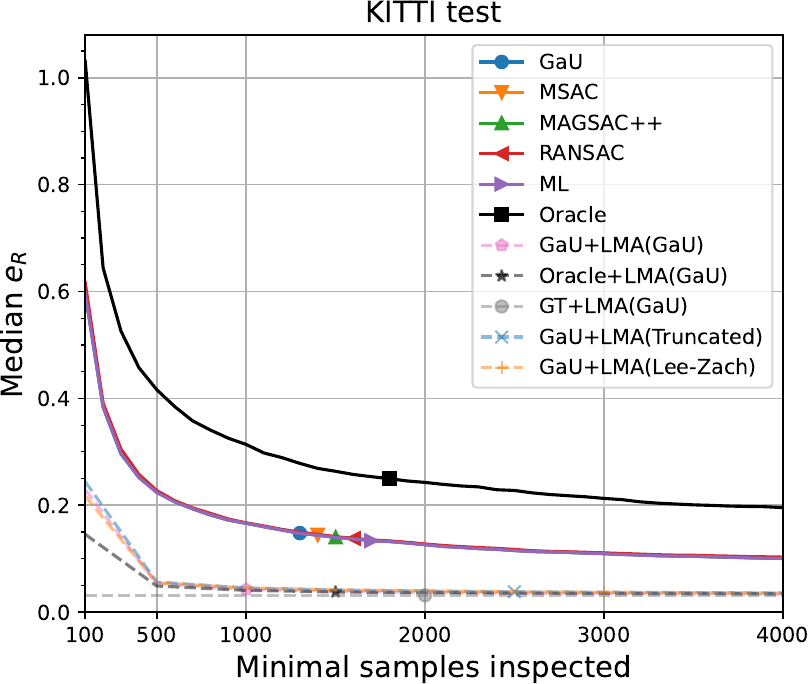}& \ \
\includegraphics[width=0.46\linewidth]{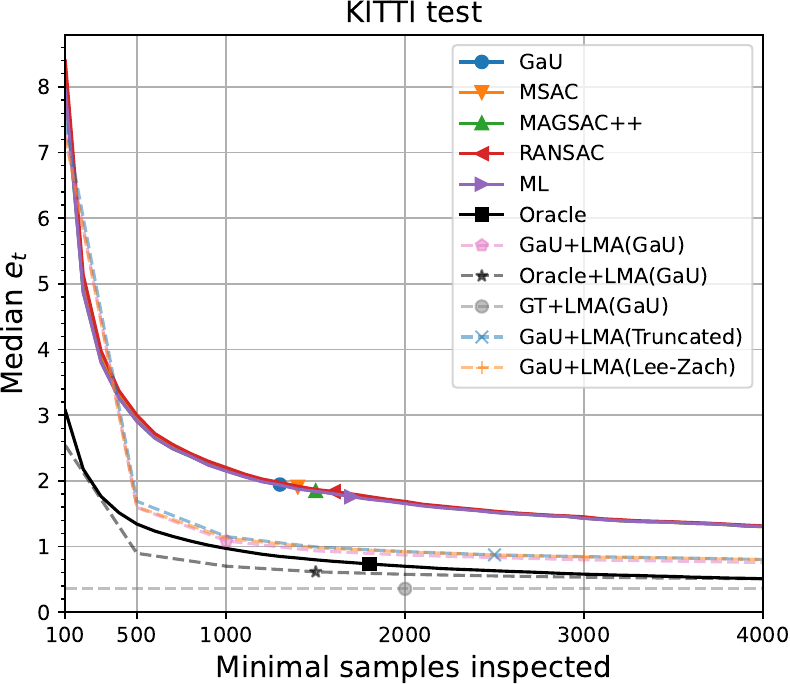}
\end{tabular}
\caption{KITTI F: Test performance versus number of samples. {\em Left:} rotation error $e_R$. {\em Right:} translation error $e_t$.
 \label{fig:KITTI-running}}

\captionof{table}{KITTI F: Test performance statistics at 4K minimal samples.
\label{tab:KITTI}
}
\resizebox{\linewidth}{!}{
\setlength{\tabcolsep}{5pt}
\begin{tabular}{llll|lll}
\toprule
\multicolumn{7}{c}{Minimal Samples Scoring}\\
\midrule
         & \multicolumn{3}{c}{mAA} & \multicolumn{3}{c}{Median} \\
Scoring: & \multicolumn{1}{c}{$e$} & \multicolumn{1}{c}{$e_R$} & \multicolumn{1}{c}{$e_t$} & \multicolumn{1}{c}{$e$} & \multicolumn{1}{c}{$e_R$} & \multicolumn{1}{c}{$e_t$} \\
\midrule
RANSAC & 0.784$\pm$0.005 & 0.980$\pm$0.001 & 0.784$\pm$0.005 & 1.31$\pm$0.03 & 0.10$\pm$0.002 & 1.31$\pm$0.03 \\
MSAC & 0.788$\pm$0.005 & 0.981$\pm$0.001 & 0.789$\pm$0.005 & 1.30$\pm$0.03 & 0.10$\pm$0.003 & 1.30$\pm$0.03 \\
MAGSAC++ & 0.790$\pm$0.005 & 0.981$\pm$0.001 & 0.790$\pm$0.005 & 1.30$\pm$0.03 & 0.10$\pm$0.003 & 1.30$\pm$0.03 \\
GaU & 0.790$\pm$0.005 & 0.981$\pm$0.001 & 0.790$\pm$0.005 & 1.31$\pm$0.033 & 0.1$\pm$0.003 & 1.31$\pm$0.03 \\
Learned & 0.790$\pm$0.005 & 0.981$\pm$0.001 & 0.790$\pm$0.005 & 1.31$\pm$0.032 & 0.1$\pm$0.003 & 1.31$\pm$0.03 \\
Oracle & 0.924$\pm$0.002 & 0.965$\pm$0.001 & 0.929$\pm$0.002 & 0.54$\pm$0.009 & 0.2$\pm$0.006 & 0.51$\pm$0.01 \\
\midrule
\multicolumn{7}{c}{Local Optimization of the Best GaU Minimal Model (Using Camera Intrinsics)}\\
LO: & \multicolumn{1}{c}{$e$} & \multicolumn{1}{c}{$e_R$} & \multicolumn{1}{c}{$e_t$} & \multicolumn{1}{c}{$e$} & \multicolumn{1}{c}{$e_R$} & \multicolumn{1}{c}{$e_t$} \\
\midrule
IRLS-LMA GaU & 0.839$\pm$0.005 & 0.991$\pm$0.001 & 0.839$\pm$0.005 & 0.76$\pm$0.02 & 0.03$\pm$0.001 & 0.76$\pm$0.02 \\
PoseLib (Truncated) & 0.834$\pm$0.005 & 0.991$\pm$0.001 & 0.834$\pm$0.005 & 0.80$\pm$0.02 & 0.04$\pm$0.001 & 0.80$\pm$0.02 \\
PoseLib (Le-Zach) & 0.835$\pm$0.005 & 0.991$\pm$0.001 & 0.835$\pm$0.005 & 0.80$\pm$0.02 & 0.04$\pm$0.001 & 0.80$\pm$0.02 \\
\bottomrule
\end{tabular}
}
\end{figure}

\paragraph{KITTI F estimation}
We repeated the evaluation for fundamental matrix estimation on KITTI dataset. Fundamental matrix estimation is very similar to the essential matrix estimation. Essentially, only the minimal solver differs. We used 7-point solver, which estimates fundamental matrix $F$. Knowing the camera intrinsics $K_1$, $K_2$, we convert it to a pseudo-essential matrix $\tilde E = K_2 \T E K_1$, where ``pseudo'' reflects the fact that $\tilde E$ does not have to obey the constraints of the essential matrix. For computing the error to the ground truth we project $\tilde E$ onto the essential matrix manifold by forcing the spectrum to be $(1,1,0)$ and then decompose the resulting $E$ into $(R,t)$ using standard methods\footnote{Both {\ttfamily kornia.geometry.decompose\_essential\_matrix} and {\ttfamily cv2.recoverPose} give the same result regardless whether $E$ or $\tilde E$ is given.} and compute the pose error.
\cref{fig:KITTI-val} shows the validation results, \cref{fig:KITTI-running} shows test running plots for rotation and translation errors. One can observe a super-oracle performance in rotation error in \cref{fig:KITTI-running}. There is no contradiction in this result: the oracle selects models that are more accurate in the total pose error $e$ but the rotation error might be higher than for the models selected using scoring methods. The breakdown of numerical results is given in~\cref{tab:KITTI}.

We have not implemented specialized reparameterization / local optimizers for $F$. Therefore, when applying LMA optimization in this experiment, we cheat: we use the knowledge of camera intrinsics (which should not be available if we consider $F$ estimation) to convert $F$ to $E$ and then optimize $E$ using its parametrization. All LMA plots in \cref{fig:KITTI-running} (resp. local optimization results in \cref{tab:KITTI}) are obtained using this method.

The experiment reconfirms that there is no difference amongst scoring methods also for 7-point minimal models for $F$ estimation. In the case of KITII dataset, even RANSAC performs equally well. The LMA-optimized solutions, including Oracle+LMA, are close to the Oracle solutions. 
This dataset has dominant forward motion, meaning small relative rotation. It also has substantially higher number of correspondences and higher inlier ratio. Minimizing the score therefore can find models closer to GT, in agreement with the score consistency in \cref{sec:score-vs-error}.

\clearpage

\end{document}